\documentclass{article}

\usepackage[final]{neurips_2025}

\usepackage[utf8]{inputenc} 
\usepackage[T1]{fontenc}    
\usepackage{url}            
\usepackage{booktabs}       
\usepackage{amsfonts}       
\usepackage{nicefrac}       
\usepackage{microtype}      
\usepackage{xcolor}         

\usepackage{enumitem}
\usepackage{amsmath}        
\definecolor{metablue}{HTML}{0064E0}
\usepackage[pagebackref,breaklinks,colorlinks,allcolors=metablue]{hyperref}
\usepackage{graphicx}       
\usepackage{caption} 
\usepackage{multirow}

\usepackage{amsmath, amsthm, amsfonts, amssymb, bbm}
\usepackage{dsfont} 
\usepackage{multirow}
\usepackage{array}
\usepackage{pifont}
\usepackage{stfloats}
\usepackage{makecell}
\usepackage{bm}
\usepackage{overpic}
\usepackage{transparent}

\DeclareMathOperator*{\argmax}{argmax}
\newcommand{\gray}[1]{\textcolor{gray}{#1}}

\usepackage{xspace}
\makeatletter
\DeclareRobustCommand\onedot{\futurelet\@let@token\@onedot}
\def\@onedot{\ifx\@let@token.\else.\null\fi\xspace}

\def\eg{\emph{e.g}\onedot} 
\def\ie{\emph{i.e}\onedot}

\def\etal{\emph{et al}\onedot}
\makeatother

\newcommand{\paragraphcustom}[1]{\vspace{0pt}\noindent\textbf{#1}}

\title{Transferable Black-Box One-Shot Forging of Watermarks via Image Preference Models}

\author{%
  Tom\'{a}\v{s} Sou\v{c}ek$^{1}$ \quad Sylvestre-Alvise Rebuffi$^{1}$ \quad Pierre Fernandez$^{1}$ \quad Nikola Jovanovi\'{c}$^{2}$\thanks{Work done during an internship at Meta.}\\ \textbf{Hady Elsahar}$^{1}$ \quad \textbf{Valeriu Lacatusu}$^{1}$ \quad \textbf{Tuan Tran}$^{1}$ \quad \textbf{Alexandre Mourachko}$^{1}$\\
  $^1$Meta FAIR \quad $^2$ETH Zurich\\
  \texttt{soucek@meta.com} \\
}

\begin{document}
\maketitle

\begin{abstract}
Recent years have seen a surge in interest in digital content watermarking techniques, driven by the proliferation of generative models and increased legal pressure. 
With an ever-growing percentage of AI-generated content available online, watermarking plays an increasingly important role in ensuring content authenticity and attribution at scale.
There have been many works assessing the robustness of watermarking to removal attacks, yet, watermark forging, the scenario when a watermark is stolen from genuine content and applied to malicious content, remains underexplored. 
In this work, we investigate watermark forging in the context of widely used post-hoc image watermarking. Our contributions are as follows. First, we introduce a preference model to assess whether an image is watermarked. 
The model is trained using a ranking loss on purely procedurally generated images without any need for real watermarks. 
Second, we demonstrate the model's capability to remove and forge watermarks by optimizing the input image through backpropagation. This technique requires only a single watermarked image and works without knowledge of the watermarking model, making our attack much simpler and more practical than attacks introduced in related work. 
Third, we evaluate our proposed method on a variety of post-hoc image watermarking models, demonstrating that our approach can effectively forge watermarks, questioning the security of current watermarking approaches.
Our code and further resources are publicly available\footnote[3]{\url{https://github.com/facebookresearch/videoseal/tree/main/wmforger}}.
\end{abstract}

\section{Introduction}

Digital watermarking is an important technology that enables content identification and attribution by embedding imperceptible signals into images without compromising their visual quality.
In recent years, the demand for such technology has increased, driven by the growth of online sharing platforms and the development of generative AI models.
It is also particularly appealing for moderation and IP protection of user-generated content at scale, where copy detection and manual verification are insufficient.
More importantly, we now also need provenance verification systems to distinguish genuine from machine-generated material, especially in settings involving misinformation, fraud, or intellectual property claims.
Regulatory frameworks such as the EU AI Act and the U.S. Executive Order on AI now explicitly call for watermarking mechanisms to ensure transparency and traceability in AI-generated content.
Among the various watermarking approaches, post-hoc watermarking---where the watermark is applied after content generation---has gained popularity for its modularity and ease of deployment, in both of these scenarios.
As a case in point, it is already integrated into many practical systems to label and trace AI-generated images or videos~\citep{deepmind2023watermarking, deepmind2024watermarking, hilbert2025watermarking}, and to protect the authenticity of photos of press agencies or taken with cameras~\citep{afp2025afp, nikon2024nikon}.

However, while the robustness of these watermarking schemes to common removal attacks has been widely investigated, their security against adversarial misuse remains rather underexplored.
In particular, watermark forging, \ie, the process by which an attacker can steal or create a counterfeit watermark and apply it to new media, can be problematic in many scenarios.
Unlike watermark removal, forging does not seek to erase a watermark, but rather to deceive downstream detection systems and create significant security vulnerabilities.
For instance, it could flood detection systems with false positives, or, in the case where watermarking is applied on authentic images, it could be used to create fake content that appears authentic.

The literature on image-based watermark attacks predominantly focuses on removal techniques, such as diffusion-based purification~\citep{zhao2024invisible, nie2022diffusion}, model inversion~\citep{liu2025image}, or access-based decoding attacks~\citep{saberi2024robustness,labiad2024log}. 
These approaches often assume extensive access to watermarked data, decoding APIs, or the watermark generation pipeline---assumptions that do not hold in realistic adversarial settings. 
Moreover, they tend to induce perceptual degradation or require computationally expensive processes. 
By contrast, forging attacks are less studied, and existing attempts rely on strong assumptions, such as paired watermarked and clean datasets~\citep{wang2021watermark} or the ability to train deep generative models on thousands of watermarked samples~\citep{dong2025imperceptible, li2023warfare}. 
These are impractical in real-world, black-box scenarios where adversaries might only observe a single watermarked image.

\begin{figure}[t]
    \centering
    \includegraphics[width=0.99\linewidth, clip, trim={0 5.85in 6in 0}]{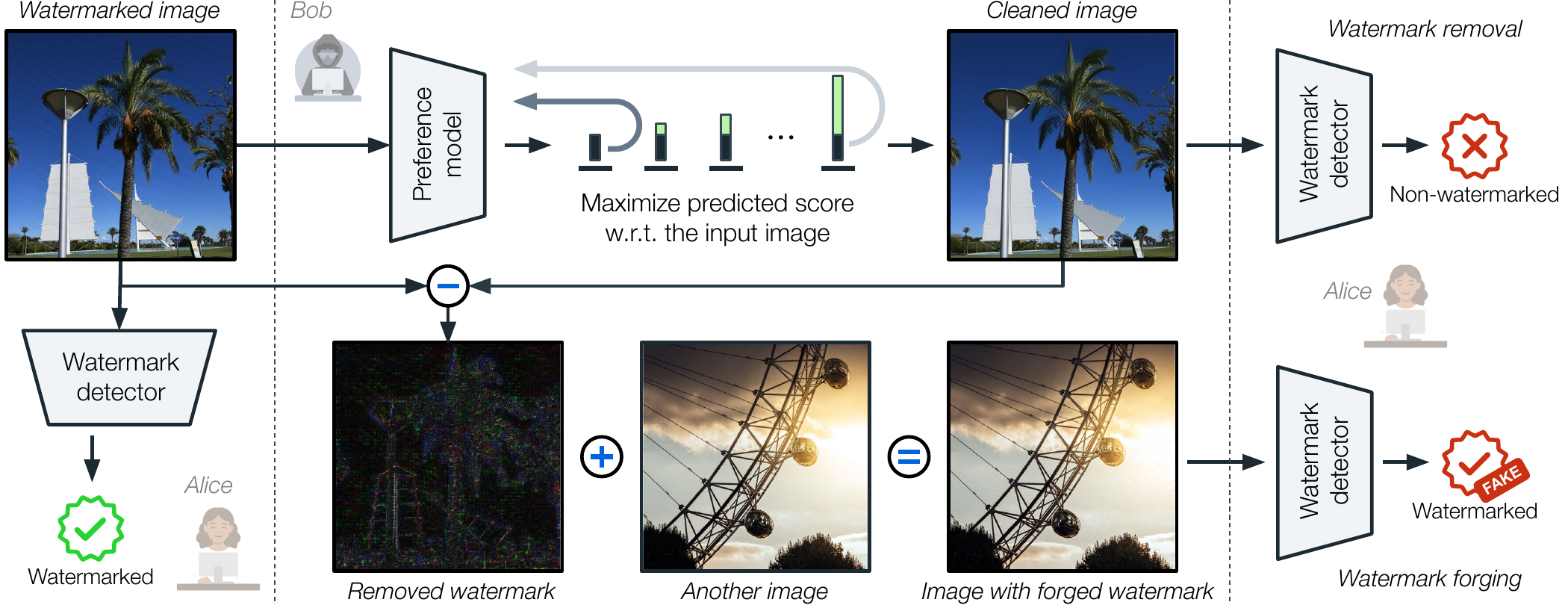}
    \caption{
    \textbf{Overview of our attack pipeline.} 
    Given a single watermarked image (left), Bob's goal is to either \emph{remove} the watermark (top path) or \emph{forge} it onto another image (bottom path). 
    He first trains a preference model to predict a score indicating the likelihood that an image will be watermarked, and then optimizes the input image to maximize this score. 
    This enables both watermark removal (producing a clean image classified as non-watermarked) and watermark forging (producing an image classified as watermarked). 
    Notably, the approach does not require access to the watermarking model or paired~data.
    }
  \label{fig:overview}
\end{figure}

In this work, we introduce a new watermark attack designed specifically for these low-resource and black-box settings. Our pipeline is depicted in Figure~\ref{fig:overview}.
We first introduce a preference model trained using a ranking-based supervision scheme on synthetically altered images.
These artifacts are generated procedurally, requiring no real watermarked data. 
The model learns to score image candidates based on how likely they are to contain a watermark, implicitly capturing the structure of ``unnatural'' artifacts. Then, the model can be used as a surrogate loss for backpropagation-based image optimization, enabling both watermark removal and watermark forging. 
Notably, this is achieved with no access to the watermarking algorithm, no paired data, and only a single example of a watermarked image.
Our design of watermark forging from a single watermarked image mirrors realistic attack settings as it allows for attacks against content-aware watermarking models---the scenario where other types of attacks often fail. To this end, our work provides a more credible assessment of post-hoc watermarking scheme vulnerabilities in the wild.
Our contributions are threefold:
\begin{itemize}[leftmargin=2em]
\item We introduce an image preference model trained solely on procedurally perturbed images using a ranking loss, which avoids reliance on actual watermarked content or decoding models.
\item We present a gradient-based attack procedure that uses the preference model to remove or forge watermarks via direct optimization on image pixels, requiring no knowledge of the original watermarking scheme.
\item We conduct comprehensive evaluations across a range of post-hoc watermarking schemes, demonstrating that our method achieves strong performance while assuming more realistic threat models.
In addition, it provides guidance on which watermarking methods are more robust against forging attacks.
\end{itemize}

\paragraphcustom{Threat model.} 
A benign actor \emph{Alice} applies watermarks to her content using a post-hoc watermarking scheme, while an adversary \emph{Bob} seeks to either remove the watermark to evade detection or forge it onto new, unrelated content to falsely pass it as authentic. 
We assume that Bob has no access to Alice’s watermarking method, paired clean-watermarked data, or any decoding interface. 
We also assume Bob gains access to a single image watermarked by Alice.
This setting reflects realistic attack scenarios where the adversary must operate in a black-box, data-scarce environment.

\section{Related work}

\noindent\textbf{Image watermarking.} A large body of work for image watermarking focuses on post-hoc watermarking~\cite{fernandez2025a}. The post-hoc watermarking, in contrast to other types of watermarking applied to the content during its generation~\cite{fernandez2023stable, wen2023tree,ci2024ringid}, has the advantage that it can also be used for attribution and authenticity verification of real content. It adds small, ideally invisible, artifacts to the image that can be later detected to verify the content's authenticity. Some works embed a single static watermark into all images~\cite{bui2023rosteals,luo2022leca}, but most works take into account the image and adjust the watermark dynamically~\cite{zhu2018hidden, luo2020distortion, ma2022towards, jia2021mbrs, bui2023trustmark, zhang2024editguard, sander2024watermark, xu2025invismark, fernandez2024video}. However, during training, the content awareness of the watermark is usually not enforced, often resulting in fairly static watermarks that can be easily forged by averaging many watermarked images~\citep{yang2024can}. In contrast, we propose a method for watermark forging that shows that even some content-aware watermarking schemes can be exploited.

\paragraphcustom{Generation-based attacks.}
A common image watermark removal attack uses diffusion models~\cite{ho2020denoising}. The approach, originally developed for preventing adversarial attacks \cite{nie2022diffusion}, introduces subtle perturbations to a watermarked image that are then removed by denoising \cite{saberi2024robustness, zhao2024invisible}. Although this approach reliably removes subtle watermarks in images, it also hallucinates new image details, deviating from the input. To counter the hallucination problem, Liu \etal \cite{liu2025image} finetune Stable Diffusion model \cite{rombach2022high} conditioned on the watermarked image for greater consistency of the purified output image with the input and Lucas \etal \cite{lukas2024leveraging} do not use diffusion at all -- they train a VAE to remove watermarks from images. On the other hand, to forge watermarks, Wang \etal \cite{wang2021watermark} train a U-Net to \textit{replicate} third-party watermarker but assume an unrealistic scenario with paired original and watermarked images. To alleviate the need for paired data, Li \etal \cite{li2023warfare} use a discriminator to train a network that adds the forged watermark to any image. Similarly, Dong \etal \cite{dong2025imperceptible} finetune a diffusion model for the same purpose. The downside of these methods is that they require thousands or tens of thousands of watermarked images with the same hidden message, which may be difficult to obtain, especially as the recent watermarking methods allow for different hidden messages based on various factors, including, for example, date or user. In contrast, our method does not require any real watermarked images and, once trained, it can be applied to any watermarked image without any adaptations to a particular watermarking method.

\paragraphcustom{Backpropagation-based attacks.} The straightforward way of evading watermark detection is to perturb the watermarked image by back-propagating through the watermark decoder and maximizing the probability of detecting a random message \cite{jiang2023evading}. However, this attack requires access to the watermark detector, which is not the case in practice.
To remove watermarks from images without access to the watermark detector, UnMarker~\cite{kassis2024unmarker} perturbs the image to maximize its difference to its original watermarked copy in the Fourier spectrum while minimizing its difference in Euclidean and LPIPS~\cite{zhang2018unreasonable} metrics. Liang \etal \cite{liang2025baseline} show that the Deep Image Prior \cite{ulyanov2018deep} framework can remove watermarks produced by modern watermarking methods, yet it is prohibitively expensive and needs model training for each watermarked image.
Saberi \etal \cite{saberi2024robustness} remove watermarks from images using a projected gradient descent on a classifier trained to distinguish between watermarked and non-watermarked images, requiring access to large collection of watermarked images. Hu \etal \cite{hu2025a} replace the need for real watermarked images by training a collection of 100 surrogate watermarking models which are then used to compute image perturbation that likely evades detection.
Lastly,~M{\"u}ller \etal \cite{muller2024black} show that semantic diffusion model-based watermarks can be both removed and forged via backpropagation through the iterative diffusion process of a surrogate diffusion model.
In contrast to these methods, we can forge watermarks of any post-hoc watermarking method, we do not need a large set of surrogate models, and we require only a single watermarked image to perform our attack.

\section{Method}

Given a single watermarked image $\bm{x}_w$, our goal is to extract a forged watermark $\hat{w}$ that can be added to any image $\bm{y}$ such that the resulting image $\bm{y}_{\hat{w}}$ is detected as watermarked by the respective watermark detector.
To produce a realistic attacked image, our second criterion is that the forged watermark $\hat{w}$ should make imperceptible modifications to the input image $\bm{y}$. Otherwise, one could artificially boost the watermark detection performance by increasing the magnitude of the forged watermark $\hat{w}$.
Besides having only one watermarked image, the main challenge of this one-shot forging task is that we do not know any prior information about the watermark $w$, and we do not have access to the watermark detector.

We propose a two-step approach illustrated in Figure~\ref{fig:overview} where (i) we estimate the watermark $\hat{w}$ from the single watermarked image $\bm{x}_w$ at our disposal and (ii) we use the estimated watermark $\hat{w}$ to forge new watermarked images $\bm{y}_{\hat{w}}=\bm{y}+\hat{w}$ for any given image $\bm{y}$. For the key step of estimating the watermark from a single image, we train an image prior model that we use to separate the artifacts produced by a watermarking model from the original content of the image. To build the image prior model, we take inspiration from the Large Language Model literature~\cite{ouyang2022training} and train a preference model $R$ using a ranking loss to prefer original images to synthetically corrupted images. Note that we do not use any watermarking model to build the preference model, only the synthetic corruptions detailed in Section~\ref{sec:method1}. Then, we can optimize the watermarked image $\bm{x}_w$ to produce a clean image~$\hat{\bm{x}}$ by maximizing the preference score of the preference model. The estimated watermark can be obtained by a simple subtraction $\hat{w}=\bm{x}_w-\hat{\bm{x}}$ and it can be used to forge new watermarked images $\bm{y}_{\hat{w}}=\bm{y}+\hat{w}$. In detail, we present the training of our preference model in Section~\ref{sec:method1}. Then, in Section~\ref{sec:method2}, we describe how to remove and forge image watermarks using the model.

\subsection{Preference model training}\label{sec:method1}
Our preference model $R$ is a ConvNeXt \cite{woo2023convnext} with an RGB image $\bm{x}$ as input. The model predicts a single score $R(\bm{x})\in\mathbb{R}$. Higher score values indicate that the input image is preferred, \ie, the image is of high quality and without any artifacts, while lower score values indicate the opposite. We detail how we train the model next.

\paragraphcustom{Preference loss.} Suppose we are given two variants $\bm{x}^+$, $\bm{x}^-$ of the image $\bm{x}$ where $\bm{x}^+$ denotes the preferred and $\bm{x}^-$ dispreferred variants of that image. For example, the two image variants can be different augmentations of the same image, where the dispreferred variant is strongly distorted. Our model $R$ is trained to predict the probability $p(\bm{x}^+ \succ \bm{x}^-)$ that the image $\bm{x}^+$ is preferred to $\bm{x}^-$. We chose the popular Bradley-Terry model~\cite{bradley1952rank} to define the preference distribution as:
\begin{equation}
    p(\bm{x}^+ \succ \bm{x}^-) = \frac{\exp\left(R(\bm{x}^+)\right)}{\exp\left(R(\bm{x}^+)\right) + \exp\left(R(\bm{x}^-)\right)}.
\end{equation}
Given a dataset of the image preference pairs ($\bm{x}^+$, $\bm{x}^-$), we can train the model $R$ using the negative log-likelihood loss:
\begin{equation}\label{eq:loss}
    -\mathbb{E}_{(\bm{x}^+, \bm{x}^-)} \left[ \log \sigma\left(R(\bm{x}^+) - R(\bm{x}^-)\right)\right].
\end{equation}

\begin{figure}[t!]
    \newcommand{\imwidth}{0.14\textwidth}

    \centering
    \scriptsize
    \setlength{\tabcolsep}{0pt}
    \begin{tabular}{c@{\hskip 2pt}c@{\hskip 20pt}c@{\hskip 2pt}c@{\hskip 20pt}c@{\hskip 2pt}c@{\hskip 0pt}}
    \multicolumn{2}{c}{\hspace{-3em}\small \textit{Wave style}} &
    \multicolumn{2}{c}{\hspace{-3em}\small\textit{Noise}} &
    \multicolumn{2}{c}{\hspace{-0em}\small\textit{Line style}} \\[2pt]

    \includegraphics[width=\imwidth]{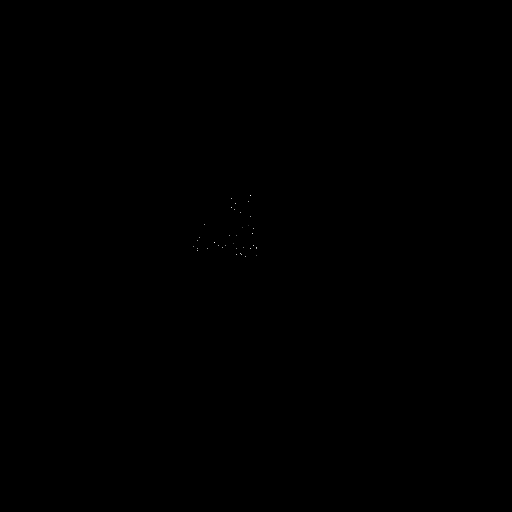} & 
    \includegraphics[width=\imwidth]{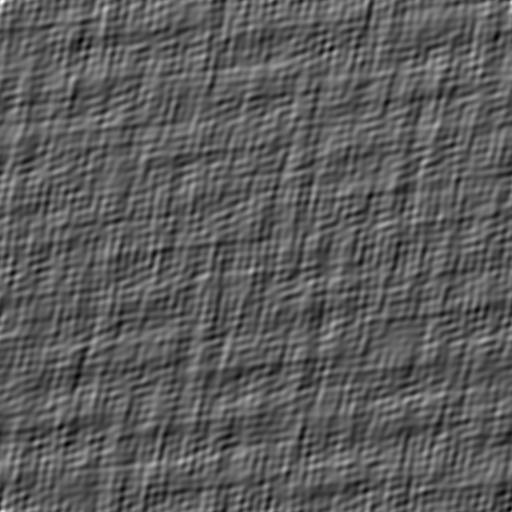} & 
    \includegraphics[width=\imwidth]{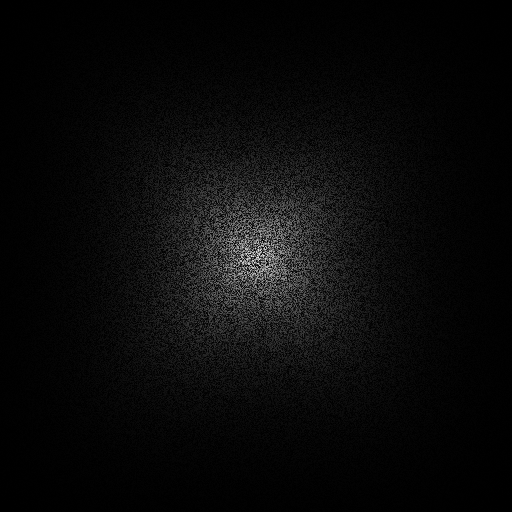} & 
    \includegraphics[width=\imwidth]{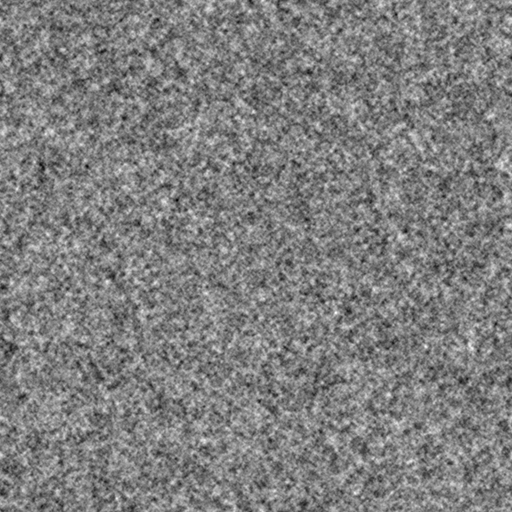} & 
    \includegraphics[width=\imwidth]{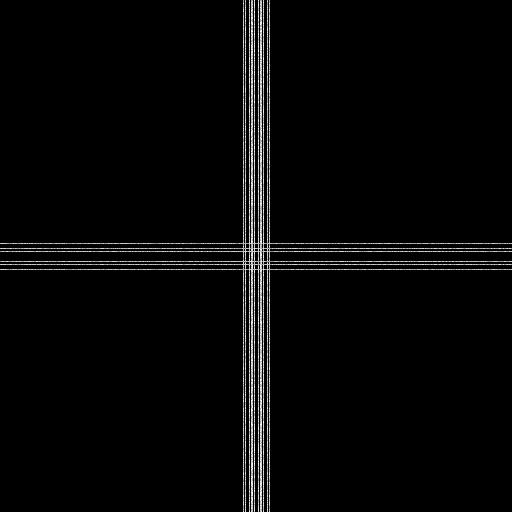} & 
    \includegraphics[width=\imwidth]{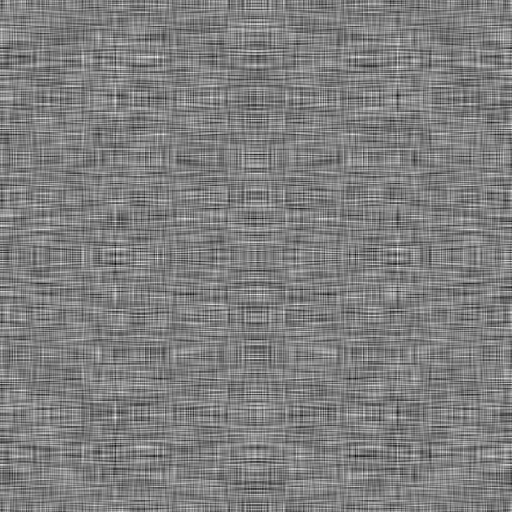} \\
    \textit{Fourier} & \textit{Image} & 
    \textit{Fourier} & \textit{Image} & 
    \textit{Fourier} & \textit{Image} \\
    \end{tabular}
    \caption{
    \textbf{Synthetic artifacts.} We train our preference model to prefer original images over synthetically corrupted images. To cover a broad range of possible artifacts, we use three types of artifacts: wave style, noise, and line style. The artifacts are first generated in the Fourier space and then transformed to the image space to be added to the original images. 
    }\label{fig:example_artifacts} 
\end{figure}

\paragraphcustom{Training data.}
Our definition of the loss function allows for any source of image preference pairs. For example, these pairs can be obtained by collecting human preferences~\cite{zhang2018unreasonable}. However, in this work, we show that artificially generated image preference pairs serve as a powerful signal for the model to learn a strong image prior. In our case, we choose the real image $\bm{x}$ as the preferred image $\bm{x}^+ = \bm{x}$. The dispreferred image is created by adding a synthetic artifact $\omega$ to the real image $\bm{x}^-=\bm{x}+\omega$. We generate these artifacts in the Fourier space by randomly choosing between wave style artifacts, line style artifacts, and noise.
In detail, suppose $\mathcal{F}(i,j)\in\mathbb{R}$ is the amplitude spectrum of the generated artifact $\omega$ of shape $H\times W$ with $i\in\{-H/2, \dots,H/2\}$, $j\in\{-W/2, \dots,W/2\}$ and the zero-frequency component centered at $(i,j)=(0,0)$. The artifacts are created as follows.
\begin{itemize}[leftmargin=2em]
    \item \textit{Wave style artifacts:} The amplitude is non-zero $\mathcal{F}(i,j)\neq0$ for $(i,j)\in \{(r^{4/5}\cos(\theta), r^{4/5}\sin(\theta))\}_{1}^N$ where $N\sim\mathcal{U}\{2,\dots, N_{max}\}$, $r\sim\mathcal{U}[0, r_{max}]$, and $\theta\sim\mathcal{U}[0,2\pi]$. The notation $\mathcal{U}[\cdot]$ or $\mathcal{U}\{\cdot\}$ denotes random sampling from a given range or set, respectively.
    
    \item \textit{Noise:} The amplitude is randomly sampled for every $(i,j)$ as $\mathcal{F}(i,j)\sim\mathcal{U}[0, \exp\left(-\|(i/\sigma^2,j/\sigma^2)\|_p\right)]$ where $\sigma^2\sim\mathcal{U}[s_{min}, s_{max}]$, and $\|\cdot\|_p$ is $p$-norm with $p=4-3\sqrt{p'}$, $p'\sim\mathcal{U}[0,1]$.
    
    \item \textit{Line style artifacts:} The amplitude is non-zero $\mathcal{F}(i,j)\neq0$ for horizontal lines $(i,j)$ such that $i$ or $-i\in\{l\}_{1}^M$ where $M\sim\mathcal{U}\{3,\dots, M_{max}\}$, $l\sim\mathcal{N}(0, \rho^2)$ and $\rho\sim\mathcal{U}[s'_{min}, s'_{max}]$. Vertical lines are sampled analogously.
\end{itemize}
The phase component of the Fourier spectrum is selected randomly, and after the transformation into the image space, the artifacts are rescaled to the range $[-0.05,0.05]$.
Examples of the generated artifacts and their generated Fourier amplitude spectra are shown in Figure~\ref{fig:example_artifacts}. To increase the artifact diversity, we randomly sample either RGB or grayscale artifacts. In addition, with 50\% probability, we randomly multiply $\omega$ by a Just Noticeable Differences (JND) map, effectively localizing the artifacts in high-frequency image regions only. 
Please see the appendix for details.

\paragraphcustom{Adversarial training.} 
The optimal preference model $R^*$ trained using Equation~\eqref{eq:loss} would predict $R^*(\bm{x}_w)=-\inf$ for the watermarked image and $R^*(\bm{x})=\inf$ for the original non-watermarked image. Therefore, one would hope that optimizing $\bm{x}_w$ to maximize the preference score $R(\bm{x}_w)$ would reduce the watermark artifacts in $\bm{x}_w$ and retrieve an image similar to the original $\bm{x}$. However, we observe in our experiments that backpropagating through the preference model trained with Equation~\eqref{eq:loss} not only fails to reduce the watermark artifacts but also adds new artifacts such as checkerboard patterns like in Figure~\ref{fig:ablqualitative}. This is analogous to what is observed in the robustness literature~\cite{santurkar2019image,tsipras2018robustness,croce2025adversarially}, where image classifiers trained without adversarial training fail to produce interpretable gradients.
Therefore, we propose to use adversarially perturbed images $\tilde{\bm{x}}^-=\bm{x}^-+\epsilon \cdot\nabla_{\bm{x}^-}R(\bm{x}^-)$ to create an extra image pair ($\bm{x}^+$, $\tilde{\bm{x}}^-$) as additional training data. The intuition behind this extra image pair is that if we add a perturbation smaller than the watermark artifacts, the resulting image should not be preferred over the original image by the preference model.

While our adversarial perturbations are similar to adversarial training~\cite{madry2017towards} - where the goal is to ensure robustness to malicious attacks that commonly do not exist in the real data - they are designed such that the preference model produces semantically interpretable gradients that can be used to estimate the watermark.

\subsection{Watermark removal and forging}\label{sec:method2}

Given a watermarked image $\bm{x}_w$, our goal is to watermark a new image $\bm{y}$ in a way that is recognized by a watermarking detector as genuinely watermarked. 
We achieve this by first estimating the watermark embedded in $\bm{x}_w$. To do so, we find the image $\hat{\bm{x}}=\bm{x}_w-\hat{w}$ that maximizes the preference model score $R(\hat{\bm{x}})$ within a certain number of optimization steps $k$. Our objective, written as a function of the watermark distortion $\delta$, is as follows:

\begin{equation}\label{eq:optim}
    \hat{w}=\argmax_{\delta} R(\bm{x}_w-\delta). 
\end{equation}

We optimize Equation~\eqref{eq:optim} using gradient ascent with a fixed number of steps $k$ and a fixed learning rate. This constrained optimization budget allows us to control the distortion produced on the original image.
Once we have extracted the estimated watermark $\hat{w}$ from the single watermarked image $\bm{x}_w$, we can paste it to any new image $\bm{y}$ to forge new watermarked images by summing the image and the watermark $\bm{y}_{\hat{w}}=\bm{y}+\hat{w}$. The whole process is illustrated in Figure~\ref{fig:overview}.

To forge watermarks in any resolution, we adopt the rescaling strategy introduced for watermarking high-resolution images~\cite{bui2023trustmark}. Given an image $\bm{x}_w$ of size $H_{ori} \times W_{ori}$, the image is first resized to a smaller resolution $H \times W$ using bilinear interpolation. Then, we estimate the watermark $\hat{w}'$ using the resized image $\bm{x}'_w$ by optimizing Equation~\eqref{eq:optim}. Then, the estimated watermark is resized to the resolution of the new image $\bm{y}$ and summed with it to form the maliciously watermarked image $\bm{y}_{\hat{w}}$:
\begin{equation}\label{eq:highres}
   \bm{y}_{\hat{w}} = \bm{y} + \text{resize}_\text{ori}(\hat{w}'),\ \ \ \ \hat{w}' = \argmax_\delta R\left( 
       \text{resize}_{H \times W}(\bm{x}_w)-\delta
   \right).
\end{equation}
The watermark removal is performed analogously by computing the watermark-free version of $\bm{x}_w$ as $\hat{\bm{x}} = \bm{x}_w - \text{resize}_\text{ori}(\hat{w}')$.

\section{Experiments}

In this section, we present the experimental setup and results of our watermark forging approach. 
First, in Section~\ref{sec:exp1}, we describe the key implementation and evaluation details. Then, in Section~\ref{sec:exp2}, we compare our approach of watermark forging and removal to related methods. Finally, in Section~\ref{sec:exp3}, we ablate the key design choices. Additional details and results are in the appendix.

\subsection{Implementation and evaluation details}\label{sec:exp1}

\noindent\textbf{Implementation details.}
We train our model, ConvNeXt V2-Tiny~\citep{woo2023convnext} on images from the SA-1b dataset~\citep{kirillov2023segment}. We resize each image to the resolution of $768 $$\times $$768$ and apply a random synthetic artifact to it. Then, both the image with and without artifact are augmented by the same random image augmentation followed by the same random crop of size $256 $$\times $$256$. 
The model is trained from scratch for 120k steps on 8 GPUs with a batch size of 16 per GPU; the training takes 60 hours using V100 GPUs. We use AdamW optimizer with a fixed learning rate of 1$\times 10^{-5}$.
In every second batch, we replace the image with the synthetic artifact by its adversarially perturbed version as described in Section~\ref{sec:method1}.
To compute the perturbation, we use two steps of gradient descent with a learning rate randomly chosen from the interval $[0.03, 0.09]$.

\paragraphcustom{Evaluation details and metrics.}
We watermark 100 images from the SA-1b validation set by all tested watermarking methods: CIN~\cite{ma2022towards}, MBRS~\cite{jia2021mbrs}, TrustMark~\cite{bui2023trustmark}, and Video Seal~\cite{fernandez2024video}. For each watermarking method, we watermark all 100 images using the same hidden message, as the methods reliant on more than a single image to remove or forge a watermark, such as Warfare and Image averaging, cannot work with multiple hidden messages.
We measure the bit accuracy of the respective watermark extractor. While the bit accuracy is dependent on the number of bits used by each method (90\% accuracy of the 32-bit CIN method effectively results in preservation of less information than 70\% accuracy of the 256-bit MBRS method), it allows for simple and interpretable comparison of different watermark removal and forging methods. For watermark removal, the bit accuracy is measured on the 100 test images with their watermarks removed. For watermark forging, we steal the watermarks from the same 100 test images and apply the stolen watermarks to a new set of 100 images. Please note that virtually any bit accuracy can be achieved by all watermark removal/forging methods if the target image is heavily edited. Therefore, we also report PSNR computed with respect to the ground truth watermark-free image. As the PSNR is very similar across different watermarking methods, we report only a single averaged number per method.

\paragraphcustom{Compared methods.} We compare with publicly available watermark forging methods and baselines. \textit{(1) Warfare} \cite{li2023warfare} is trained to add watermarks to images using a discriminator loss that distinguishes between watermarked and non-watermarked images. We train a separate model for each watermarking method for 6 days on 8 V100 GPUs with 1000 watermarked and non-watermarked images. \textit{(2) Image averaging} is a simple method proposed by Yang \etal \cite{yang2024can} that recovers the watermark by averaging and subtracting sets of watermarked and non-watermarked images. The recovered watermark can then be pasted onto a new image. In our experiments, we average 100 images. \textit{(3) Noise blending} \cite{saberi2024robustness} pastes a watermarked random noise onto a new image. \textit{(4) Gray image blending} extracts a watermark from a uniform gray image. Both Noise blending and Gray image blending serve as baselines only since they require access to a watermarking API, making them impractical in real-world scenarios.
Additionally, we also evaluate common watermark removal methods. \textit{(5) DiffPure} \cite{nie2022diffusion} is a technique that adds Gaussian noise to a watermarked image and denoises the image using a diffusion model. We use \texttt{FLUX.1 [dev]} \cite{flux2024} model and denoise for the last 3 timesteps of the default scheduler. We also adapt this method to watermark forging by pasting the residual into new images. \textit{(6) CtrlRegen}~\cite{liu2025image} finetunes a diffusion model specifically to remove watermarks. In our setup, we use their model with $\mathrm{step}=0.1$ to preserve more information from the input image. Lastly, as a baseline, we use \textit{(7)~VAE} encoding and decoding as suggested in the literature~\cite{fernandez2023stable, an2024waves, zhao2024invisible} to show how robust the watermarking methods are to neural compression.

\begin{table}
  \caption{\textbf{Watermark forging results.} Our approach outperforms the prior works in different~watermark forging scenarios, requiring only a single watermarked image with no access to the original watermarking model. While methods such as Image averaging and Warfare are competitive, they require 100s of watermarked images with the same hidden message, which may be difficult to obtain.}
  \label{tab:forging}
  \centering

{\footnotesize
\begin{tabular}{lccccc}
\toprule
\multirow{2}{*}{\vspace{-0.7em}Method} & {{CIN}} & {{MBRS}} & {{TrustMark}} & {{Video Seal}} & \multirow{2}{*}{\vspace{-0.7em}PSNR\scriptsize{($\uparrow$)}} \\
\cmidrule(lr){2-2} \cmidrule(lr){3-3} \cmidrule(lr){4-4} \cmidrule(lr){5-5}
& {Bit~acc.~\scriptsize{($\uparrow$)}} & {Bit~acc.~\scriptsize{($\uparrow$)}}& {Bit~acc.~\scriptsize{($\uparrow$)}} & {Bit~acc.~\scriptsize{($\uparrow$)}} &  \\
\midrule
\gray{Gray image blending} & \gray{1.00} & \gray{0.80} & \gray{0.54} & \gray{0.83} & \gray{52.9} \\
\gray{Noise blending ($\alpha =0.1$) \cite{saberi2024robustness}} & \gray{0.82} & \gray{0.61} & \gray{0.52} & \gray{0.53} & \gray{21.8} \\
\midrule
Warfare ($n=1000$) \cite{li2023warfare} & 0.93 & 0.50 & 0.53 & 0.74 & \textbf{39.6} \\
DiffPure (\texttt{FLUX.1 [dev]}) \cite{nie2022diffusion} & \textbf{1.00} & 0.83 & 0.59  & 0.75 & 26.6 \\
Image averaging ($n=100$) \cite{yang2024can} & \textbf{1.00} & \textbf{0.91} & \textbf{0.61} & 0.59 & 26.2 \\
\midrule
\textbf{Ours ($n=1$)} & \textbf{1.00} & 0.83 & \textbf{0.61} & \textbf{0.83} & 31.3\\
\bottomrule
\end{tabular}
}

\end{table}

\begin{table}
  \caption{\textbf{Watermark removal results.} Our approach remains competitive with related works while not suffering from the hallucination of details and textures that are present in diffusion-based methods.}
  \label{tab:removal}
  \centering

{\small
\begin{tabular}{lccccc}
\toprule
\multirow{2}{*}{\vspace{-0.7em}Method} & {{CIN}} & {{MBRS}} & {{TrustMark}} & {{Video Seal}} & \multirow{2}{*}{\vspace{-0.7em}PSNR\scriptsize{($\uparrow$)}} \\
\cmidrule(lr){2-2} \cmidrule(lr){3-3} \cmidrule(lr){4-4} \cmidrule(lr){5-5}
& {Bit~acc.~\scriptsize{($\downarrow$)}} & {Bit~acc.~\scriptsize{($\downarrow$)}}& {Bit~acc.~\scriptsize{($\downarrow$)}} & {Bit~acc.~\scriptsize{($\downarrow$)}} &  \\
\midrule
Image averaging ($n=100$) \cite{yang2024can} & \textbf{0.39} & \textbf{0.50} & 0.60 & 0.89 & 26.3 \\
\midrule
VAE (\texttt{FLUX.1 [dev]}) & 1.00 & 0.99 & 1.00 & 0.99 & \textbf{34.3} \\
DiffPure (\texttt{FLUX.1 [dev]}) \cite{nie2022diffusion} & 0.86 & 0.56 & \textbf{0.56}  & 0.60 & 25.4 \\
CtrlRegen ($\mathrm{step}=0.1$) \cite{liu2025image} & 0.86 & 0.70 & \textbf{0.56} & 0.55 & 24.4 \\
\midrule
\textbf{Ours ($n=1$)} & 0.82 & 0.64 & 0.60 & \textbf{0.49} & 31.2 \\
\bottomrule
\end{tabular}
}

\end{table}

\subsection{Comparison with the state-of-the-art}\label{sec:exp2}

\noindent\textbf{Watermark forging.} We evaluate watermark forging methods on the test set in Table~\ref{tab:forging}. We observe strong performance (high bit accuracy) of the image averaging baseline in forging CIN~\cite{ma2022towards} and MBRS~\cite{jia2021mbrs} watermarks. This can be explained by the fact that these methods produce watermarks that are generally independent of the image content. On the other hand, for Video Seal~\cite{fernandez2024video}, the image averaging approach fails as Video Seal watermarks are highly dependent on the input image. This is where our method, which can extract the watermark from a single image, significantly outperforms all the related works. Lastly, we can see that TrustMark~\cite{bui2023trustmark} is very difficult to forge, possibly due to the fact that both embedder's and decoder's outputs are greatly dependent on the input image. In the case of Video Seal, the embedder's output is highly dependent on the input image, but the decoder tends to ignore the image, making watermark forging easier.

\paragraphcustom{Watermark removal.} We also evaluate performance in watermark removal. In this case, in contrast to watermark forging, the goal is to remove the watermark; therefore, ideally, produce random bit accuracy. As shown in Table~\ref{tab:removal}, our method is highly competitive with the related works, producing high-quality images (high PSNR) while removing most of the watermark information present in the image (low bit accuracy). Similarly to watermark forging, we can see that image averaging performs well for CIN and MBRS watermarks due to their practical independence on the input image. Also, we can see that CIN watermarks are very difficult to remove -- this is likely due to the fact that CIN watermarks significantly alter the images, are very visible, and are fairly redundant due to the low bit count of the CIN method.

\begin{figure}[t!]
    \newcommand{\imwidth}{0.1928\textwidth}

    \centering
    \scriptsize
    \setlength{\tabcolsep}{0pt}
    \begin{tabular}{c@{\hskip 2pt}c@{\hskip 2pt}c@{\hskip 2pt}c@{\hskip 2pt}c@{\hskip 2pt}c}
    & \textit{actual watermark} & \textbf{Ours} & Warfare & DiffPure & Image averaging \\

    \rotatebox{90}{\hspace{1.15cm}CIN} &
    \includegraphics[width=\imwidth]{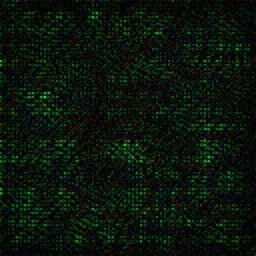} & 
    \includegraphics[width=\imwidth]{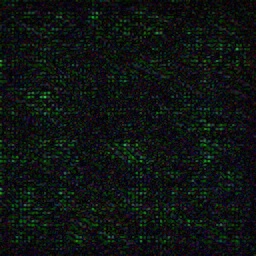} & 
    \includegraphics[width=\imwidth]{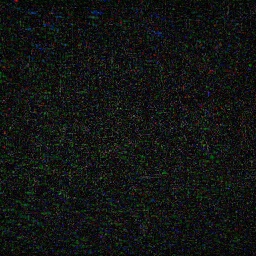} & 
    \includegraphics[width=\imwidth]{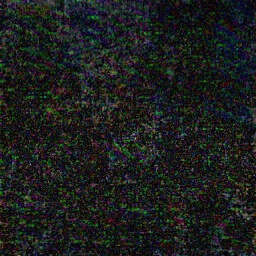} & 
    \includegraphics[width=\imwidth]{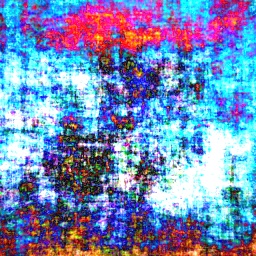} \\
    \rotatebox{90}{\hspace{0.85cm}TrustMark} &
    \includegraphics[width=\imwidth]{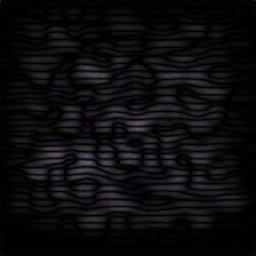} & 
    \includegraphics[width=\imwidth]{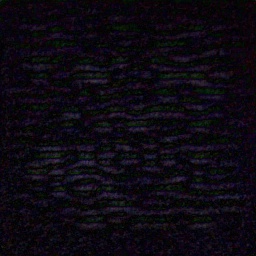} & 
    \includegraphics[width=\imwidth]{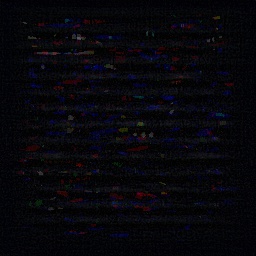} & 
    \includegraphics[width=\imwidth]{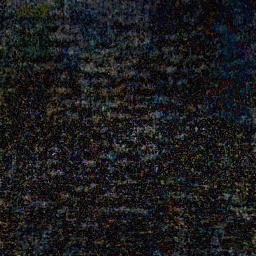} & 
    \includegraphics[width=\imwidth]{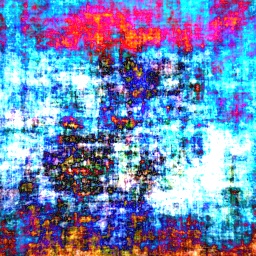} \\
    \end{tabular}
    \vspace{-0.3em}
    \caption{
    \textbf{Comparison of the forged watermarks} by different watermark forging methods. The shown watermarks are averaged from 100 different images to remove any image-specific artifacts. The recovered watermark by our method closely resembles the actual watermark (left column), containing the least number of other distracting artifacts, such as the noise present in DiffPure watermarks.
    }\label{fig:forging_avg} 
\end{figure}

\begin{figure}[t!]
    \newcommand{\imwidth}{0.192\textwidth}

    \centering
    \scriptsize
    \setlength{\tabcolsep}{0pt}
    \begin{tabular}{c@{\hskip 2pt}c@{\hskip 2pt}c@{\hskip 2pt}c@{\hskip 2pt}c@{\hskip 2pt}c}
    & \textit{clean image} & CIN & MBRS & TrustMark & Video Seal \\
    \cmidrule{2-6}

    \rotatebox{90}{\hspace{1.35cm}\textit{source image}} &
    \includegraphics[width=\imwidth]{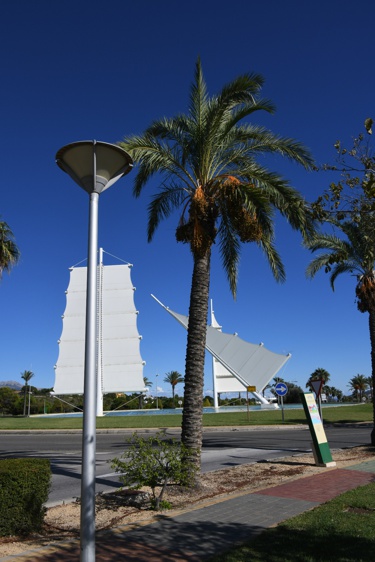} & 
    \includegraphics[width=\imwidth]{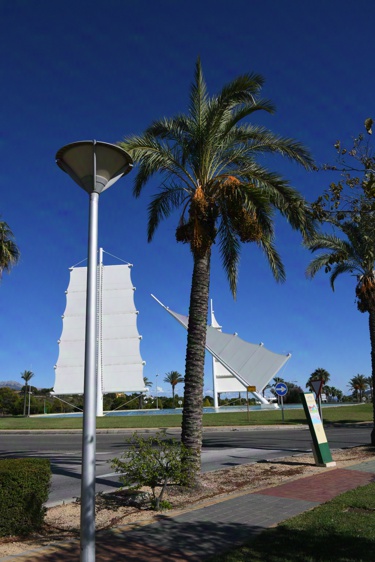} & 
    \includegraphics[width=\imwidth]{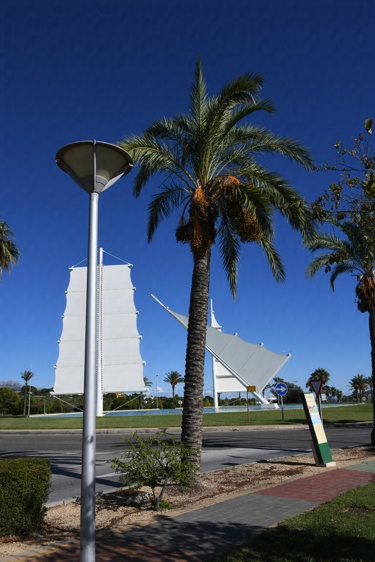} & 
    \includegraphics[width=\imwidth]{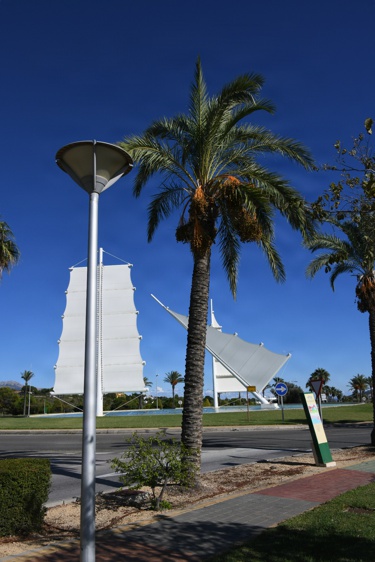} & 
    \includegraphics[width=\imwidth]{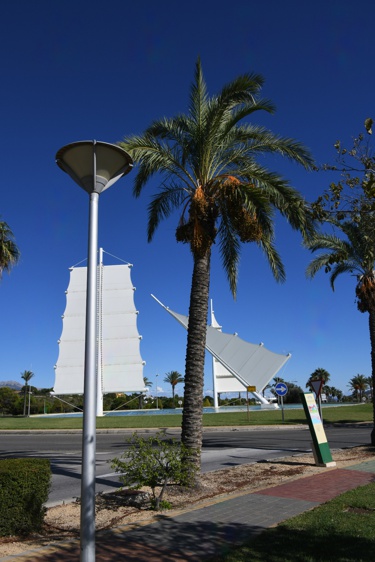} \\
    \rotatebox{90}{\hspace{0.95cm}\textit{extracted watermark}} &
    & 
    \includegraphics[width=\imwidth]{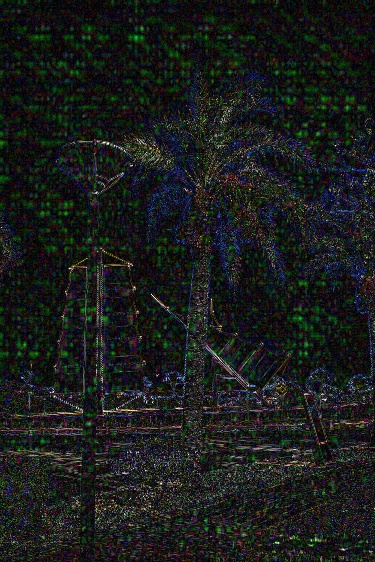} & 
    \includegraphics[width=\imwidth]{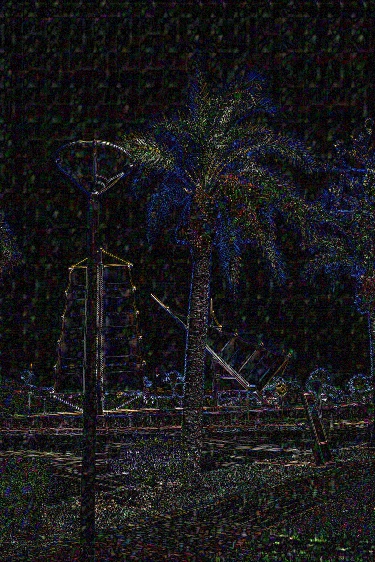} & 
    \includegraphics[width=\imwidth]{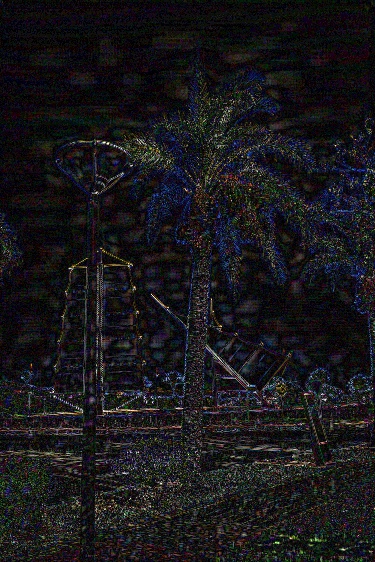} & 
    \includegraphics[width=\imwidth]{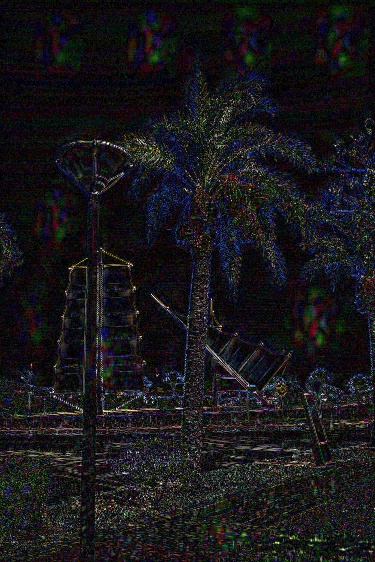} \\

    \rotatebox{90}{\hspace{1.3cm}\textit{forged image}} &
    \includegraphics[width=\imwidth]{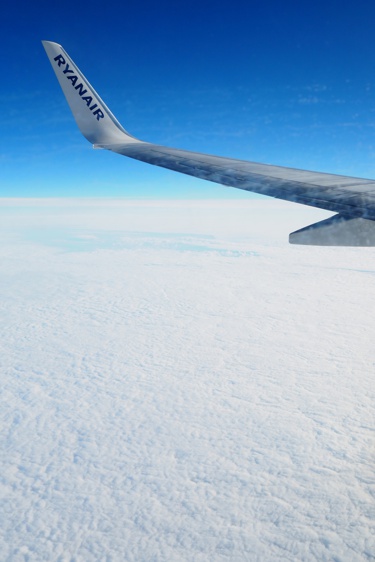} & 
    \includegraphics[width=\imwidth]{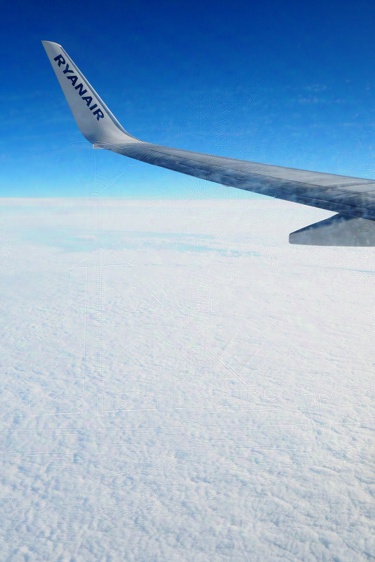} & 
    \includegraphics[width=\imwidth]{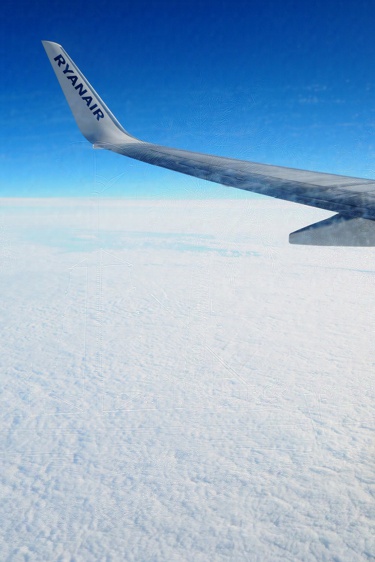} & 
    \includegraphics[width=\imwidth]{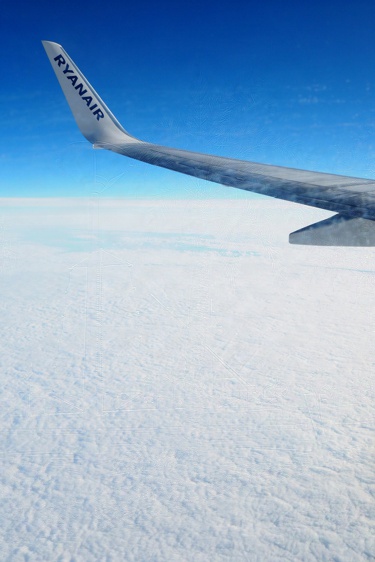} & 
    \includegraphics[width=\imwidth]{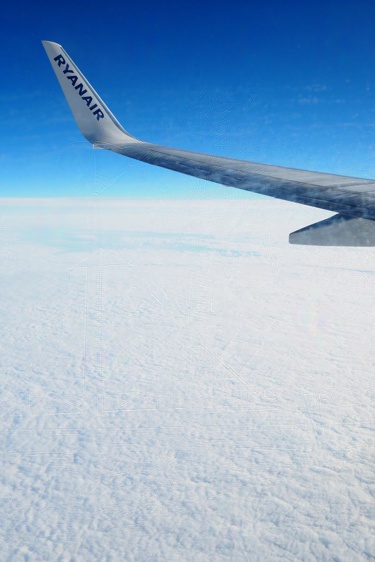} \\

    \end{tabular}
    \vspace{-0.3em}
    \caption{
    \textbf{Qualitative results for watermark forging.} The figure shows a watermarked image (top) with its watermark removed by our method (middle row). The watermark is pasted onto a new image (bottom row). We use $k=50$ steps for the watermark extraction.
    }\label{supmat:fig:wm_forging_example} 
\end{figure}

\begin{figure}[t!]
    \newcommand{\imwidth}{0.196\textwidth}

    \centering
    \scriptsize
    \setlength{\tabcolsep}{0pt}
    \begin{tabular}{c@{\hskip 2pt}c@{\hskip 2pt}c@{\hskip 2pt}c@{\hskip 2pt}c}
    \textit{watermarked image} & Image averaging & DiffPure & CtrlRegen & \textbf{Ours} \\

    \includegraphics[width=\imwidth]{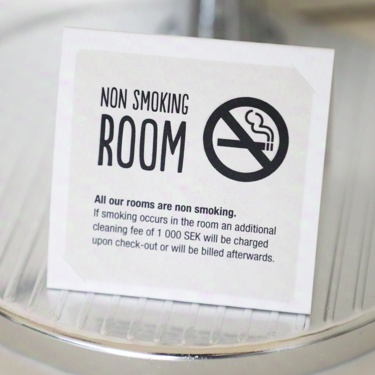} & 
    \includegraphics[width=\imwidth]{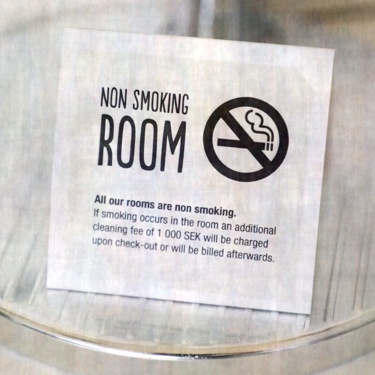} & 
    \includegraphics[width=\imwidth]{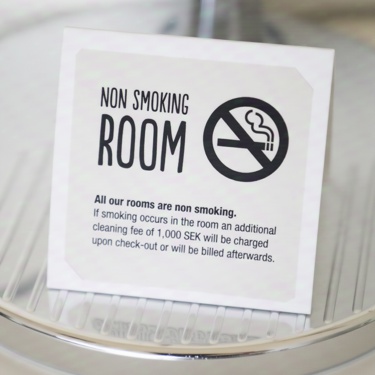} & 
    \includegraphics[width=\imwidth]{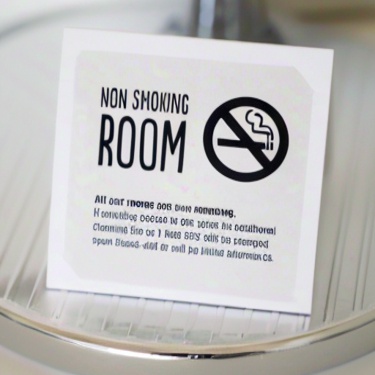} & 
    \includegraphics[width=\imwidth]{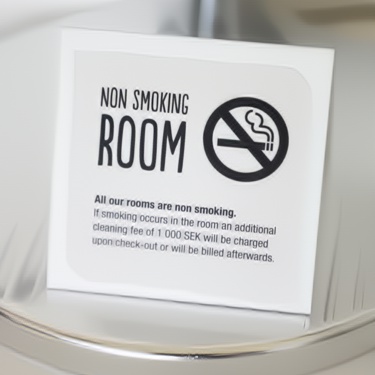} \\
    \begin{overpic}[width=\imwidth]{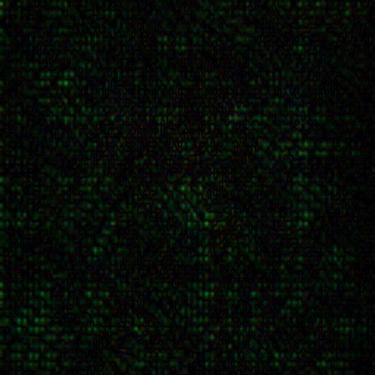}
        \put(18.5,5){\textcolor{white}{\textit{actual watermark}}}
    \end{overpic} & 
    \includegraphics[width=\imwidth]{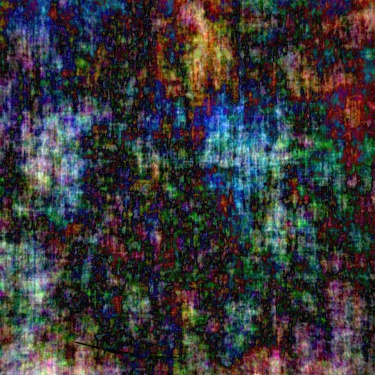} & 
    \includegraphics[width=\imwidth]{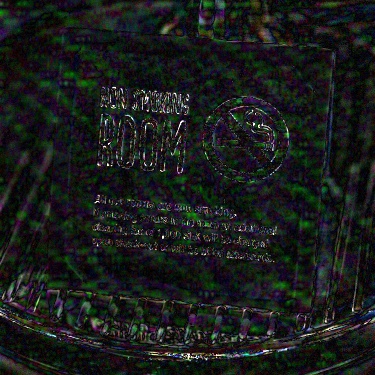} & 
    \includegraphics[width=\imwidth]{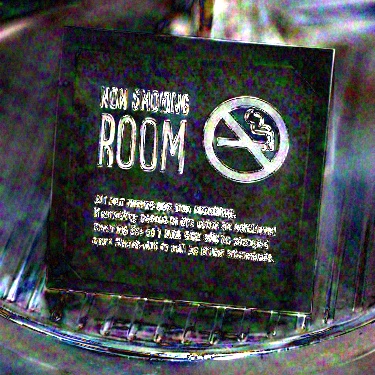} & 
    \includegraphics[width=\imwidth]{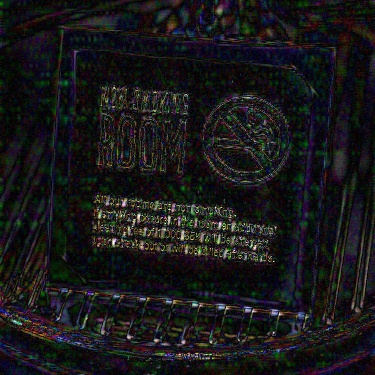} \\
    \end{tabular}
    \caption{
    \textbf{Qualitative results for watermark removal.} The figure shows a watermarked image (left) with its watermark removed by different methods (top row) and the actual removed watermark by each method (bottom row).
    }\label{fig:removal_example} 
\end{figure}

\paragraphcustom{Qualitative results.}
Results in Figures~\ref{fig:forging_avg}, \ref{supmat:fig:wm_forging_example}, \ref{fig:removal_example}, and in the appendix demonstrate the key strengths of our approach. First, in Figure~\ref{fig:forging_avg}, we present the forged watermarks obtained by various methods, averaged over 100 images to eliminate any image-specific artifacts. We can see that our method can reconstruct the watermark faithfully without requiring any per-method tuning or training, which is not the case for the Warfare method. In Figure~\ref{supmat:fig:wm_forging_example}, we show extracted watermarks forged into a new image. Lastly, in Figure~\ref{fig:removal_example}, we present the result of watermark removal, along with the removed watermark itself. While related methods, such as DiffPure and CtrlRegen, modify the input image and add new hallucinated details, our method removes only high-frequency artifacts from the image.

\begin{table}
  \caption{\textbf{Watermark forging ablation.} We test multiple variants of our model to show the effect of individual design decisions. We use the same hyperparameters for the forging process (number of steps, learning rate) for all model variants. The higher PSNR of the variants is caused by less informative gradients that do not remove the watermark artifacts effectively.}
  \label{tab:forging_ablation}
  \centering

{\footnotesize
\begin{tabular}{c@{~~}lccccc}
\toprule
& \multirow{2}{*}{\vspace{-0.7em}Method} & {{CIN}} & {{MBRS}} & {{TrustMark}} & {{Video Seal}} & \multirow{2}{*}{\vspace{-0.7em}PSNR\scriptsize{($\uparrow$)}} \\
\cmidrule(lr){3-3} \cmidrule(lr){4-4} \cmidrule(lr){5-5} \cmidrule(lr){6-6} 
& & {Bit~acc.~\scriptsize{($\uparrow$)}} & {Bit~acc.~\scriptsize{($\uparrow$)}}& {Bit~acc.~\scriptsize{($\uparrow$)}} & {Bit~acc.~\scriptsize{($\uparrow$)}} &  \\
\midrule
(1) & Binary cross-entropy loss & 0.60 & 0.53 & 0.52 & 0.47 & 39.9 \\
(2) & Hinge loss & 0.62 & 0.55 & 0.52 & 0.47 & \textbf{44.1} \\
(3) & Without perturbation & 0.97 & 0.65 & 0.52 & 0.49 & 34.7 \\
(4) & Real watermarks as training data & \textbf{1.00} & 0.67 & 0.58 & 0.77 & 36.9 \\
\midrule
(5) & \textbf{Ours} & \textbf{1.00} & \textbf{0.83} & \textbf{0.61} & \textbf{0.83} & 31.3\\
\bottomrule
\end{tabular}
}

\end{table}

\subsection{Ablations}\label{sec:exp3}

We evaluate the key design decisions of our proposed method, \ie, the use of ranking loss, input perturbation, and the artifacts used during training. Lastly, in the appendix, we also evaluate how the optimization hyperparameters affect the watermark removal.

\paragraphcustom{Training loss.} We train our model using the ranking loss as specified in Equation~\eqref{eq:loss}, but we also investigate two additional losses. We consider binary cross-entropy, where the two classes are the positive $\bm{x}^+$ and negative $\bm{x}^-$ examples, and hinge loss, often used in discriminator training~\cite{rombach2022high}.
We show in Table~\ref{tab:forging_ablation} that our decision to use the ranking loss (line 5) is crucial to achieving superior performance. The poor, almost random, performance of the model trained with binary cross-entropy (line 1) can be explained by the fact that the images $\bm{x}^+$ and $\bm{x}^-$ are very similar, and there exists no global decision threshold between the classes.

\begin{figure}[t!]
    \newcommand{\imwidth}{0.196\textwidth}

    \centering
    \scriptsize
    \setlength{\tabcolsep}{0pt}
    \begin{tabular}{c@{\hskip 2pt}c@{\hskip 2pt}c@{\hskip 2pt}c@{\hskip 2pt}c}
    \textit{Watermarked image} & (1) BCE loss & (2) Hinge loss & (3) Without perturbation & \textbf{Ours} \\

    \includegraphics[width=\imwidth]{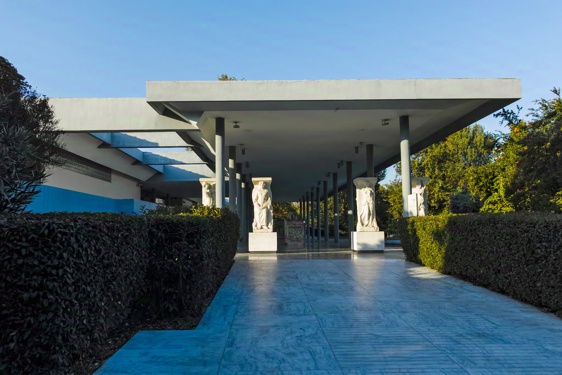} & 
    \includegraphics[width=\imwidth]{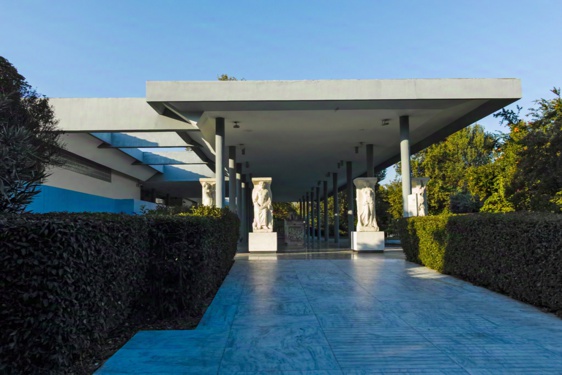} & 
    \includegraphics[width=\imwidth]{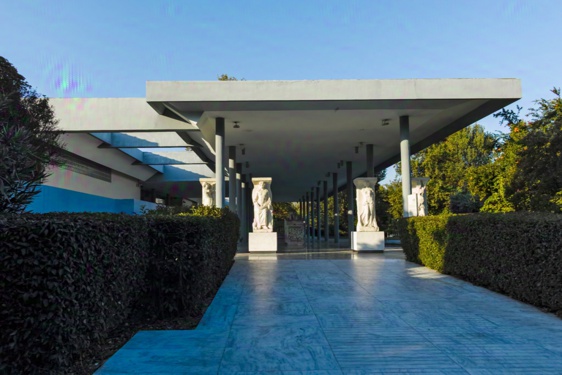} &
    \includegraphics[width=\imwidth]{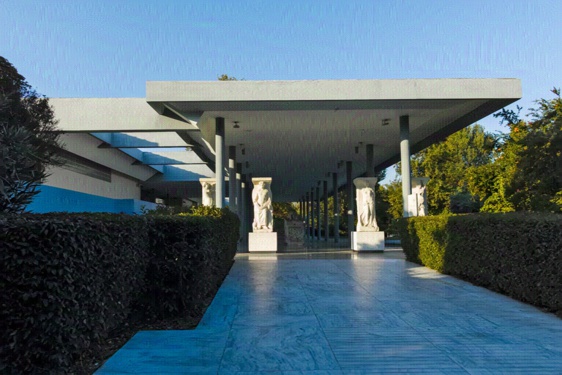} & 
    \includegraphics[width=\imwidth]{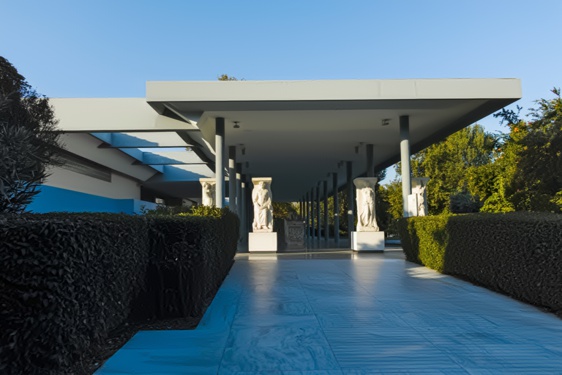} 
    \end{tabular}
    \caption{
    \textbf{Watermark removal with tested model variants.} Training our model without adversarial perturbation or with a different loss results in uninterpretable gradients and various artifacts.
    }\label{fig:ablqualitative} 
\end{figure}

\paragraphcustom{Adversarial perturbation.} During training, as explained in Section~\ref{sec:method1}, we perturb the negative samples $\bm{x}^-$ in the direction of the gradient towards the positive sample. Our perturbation is designed to create different, yet plausible negative samples, making our model more robust to different artifacts. As shown in Table~\ref{tab:forging_ablation}, lines 3 and 5, our approach, which utilizes perturbation, significantly improves upon the baseline with no perturbation used. Additionally, the gradients of the baseline without perturbation with respect to the input image are less meaningful compared to our model, whose gradients clearly point to a cleaner, less noisy image (Figure~\ref{fig:ablqualitative}, columns 2 and 3).

\paragraphcustom{Black-box vs. gray-box attack.} We train our preference model with procedurally generated artifacts, as described in Section~\ref{sec:method1}. We also test the gray-box scenario where we have access to both the original and watermarked images. Therefore, we train our preference model with the actual outputs of the watermarking methods CIN, MBRS, TrustMark, and Video Seal as the negative samples $\bm{x}^-$. As you can see from Table~\ref{tab:forging_ablation}, training on the real watermarks (line 4) performs significantly worse than training on our procedurally generated artifacts (line 5). We observe that the real watermarks are not as diverse as our generated artifacts, leading the model to overfit to the training data.

\paragraphcustom{Synthetic artifact types.} We investigate the contributions of different synthetic artifacts to the performance of the final model by training the model independently for each style of watermark pattern and evaluating its ability to remove watermarks. The results in Table~\ref{tab:synthetic_ablation} show that the model trained with the wave pattern is the most effective in removing the watermarks. However, on average, the model trained using the combination of the three watermark types produces better results.

\begin{table}
  \caption{\textbf{The effect of different synthetic artifacts on the watermark removal results.} Out of the tested artifacts, the model trained with the wave pattern is the most effective in removing the watermarks.}
  \label{tab:synthetic_ablation}
  \centering

{\footnotesize
\begin{tabular}{lcccc}
\toprule
\multirow{2}{*}{\vspace{-0.7em}Artifact type} & {{CIN}} & {{MBRS}} & {{TrustMark}} & {{Video Seal}} \\
\cmidrule(lr){2-2} \cmidrule(lr){3-3} \cmidrule(lr){4-4} \cmidrule(lr){5-5}
& {Bit~acc.~\scriptsize{($\downarrow$)}} & {Bit~acc.~\scriptsize{($\downarrow$)}}& {Bit~acc.~\scriptsize{($\downarrow$)}} & {Bit~acc.~\scriptsize{($\downarrow$)}}  \\
\midrule
Only wave style & 0.98 & 0.83 & \textbf{0.55} & \textbf{0.45} \\
Only noise      & 1.00 & 0.97 & 0.99 & 0.97 \\
Only line style & 1.00 & 0.95 & 0.99 & 0.95 \\
\midrule
All (\textbf{Ours}) & \textbf{0.82} & \textbf{0.64} & {0.60} & {0.49} \\
\bottomrule
\end{tabular}
}

\end{table}

\section{Conclusion}
We present a method capable of forging watermarks from a single watermarked image. Given its fast speed and reliance on a single watermarked image, the presented method is a more practical attack on many contemporary post-hoc watermarking techniques than related work. While some content-aware watermarking methods are fairly immune to this type of attack, we show that for other content-aware methods, the watermark can be easily stolen, questioning the security of the current post-hoc watermarking research landscape.

\paragraphcustom{Limitations.}
Our type of forging attack targets post-hoc watermarking methods. Semantic watermarking, such as Tree-Ring~\cite{wen2023tree} or RingID~\cite{ci2024ringid}, watermarks AI-generated content by altering the objects and their locations in a generated image. Our method cannot semantically change these objects; therefore, different methods must be used for forging these semantic watermarks (\eg, \cite{muller2024black}). Also, as shown in the appendix, our method may blur areas with natural high-frequency texture, such as water surfaces. Nonetheless, this issue can be partially mitigated through improved preference model training by introducing image blur as a new synthetic artifact type.

\paragraphcustom{Broader impact and safeguards.} The presented work highlights a flaw of many existing post-hoc watermarking methods and introduces a potential attack to easily forge watermarks. The attack is based on the observation that many watermark decoders are not content-aware and will accept a watermark from a different source image. To mitigate this issue, we recommend ensuring the decoder is truly content-aware, \eg, by explicitly training the decoder to reject watermarks from different source image. We believe that our insights will help to strengthen current and future watermarking techniques and contribute towards making AI safe and responsible.

\clearpage

{
\small
\bibliographystyle{ieeenat_fullname}

\bibliography{references}
}



\clearpage
\appendix

\section*{Appendix}
\section{Additional results}

\begin{figure}[t!]
    \newcommand{\imwidth}{0.192\textwidth}

    \centering
    \scriptsize
    \setlength{\tabcolsep}{0pt}
    \begin{tabular}{c@{\hskip 2pt}c@{\hskip 2pt}c@{\hskip 2pt}c@{\hskip 2pt}c@{\hskip 2pt}c}
    & \textit{clean image} & CIN & MBRS & TrustMark & Video Seal \\
    \cmidrule{2-6}

    \rotatebox{90}{\hspace{0.7cm}\textit{source image}} &
    \includegraphics[width=\imwidth]{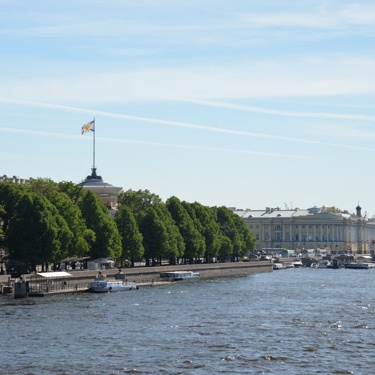} & 
    \includegraphics[width=\imwidth]{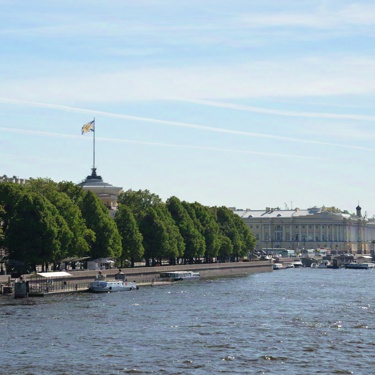} & 
    \includegraphics[width=\imwidth]{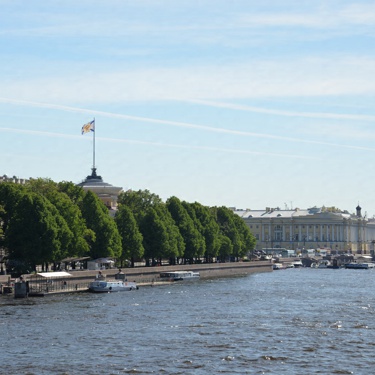} & 
    \includegraphics[width=\imwidth]{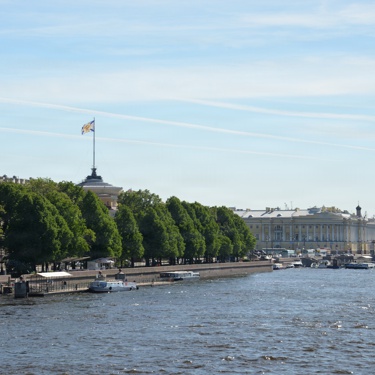} & 
    \includegraphics[width=\imwidth]{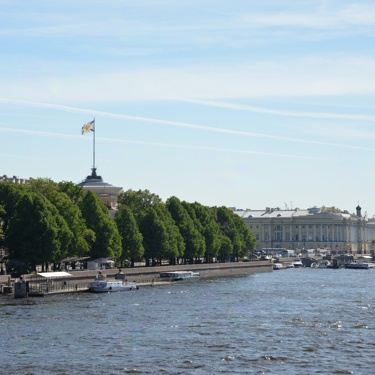} \\
    \rotatebox{90}{\hspace{0.3cm}\textit{extracted watermark}} &
    & 
    \includegraphics[width=\imwidth]{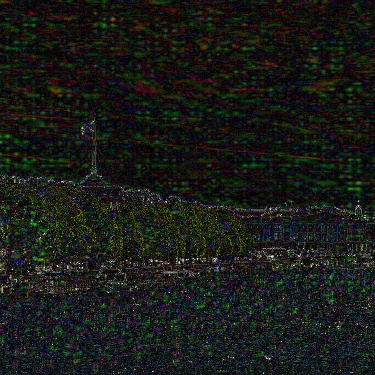} & 
    \includegraphics[width=\imwidth]{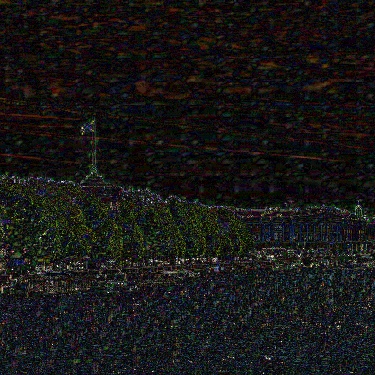} & 
    \includegraphics[width=\imwidth]{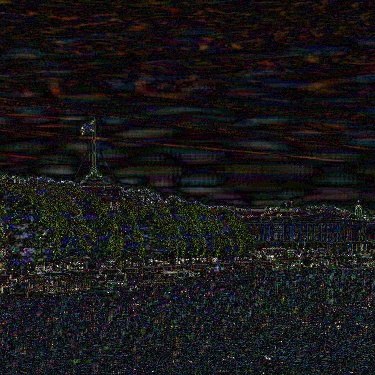} & 
    \includegraphics[width=\imwidth]{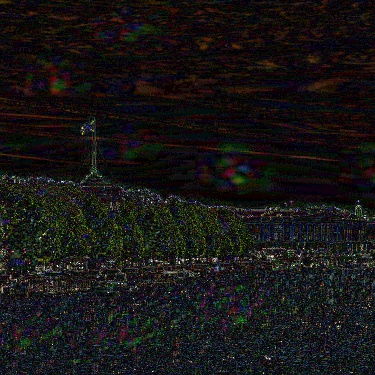} \\

    \rotatebox{90}{\hspace{0.7cm}\textit{forged image}} &
    \includegraphics[width=\imwidth]{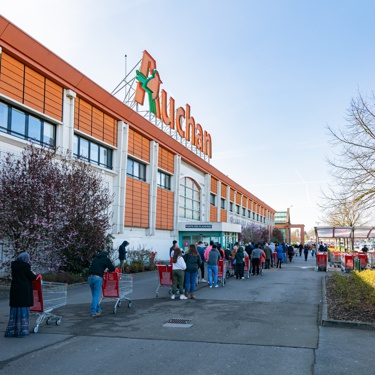} & 
    \includegraphics[width=\imwidth]{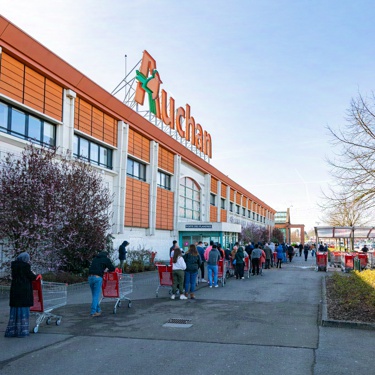} & 
    \includegraphics[width=\imwidth]{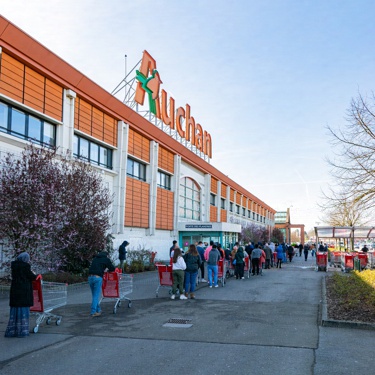} & 
    \includegraphics[width=\imwidth]{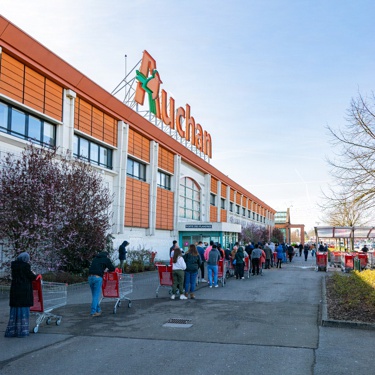} & 
    \includegraphics[width=\imwidth]{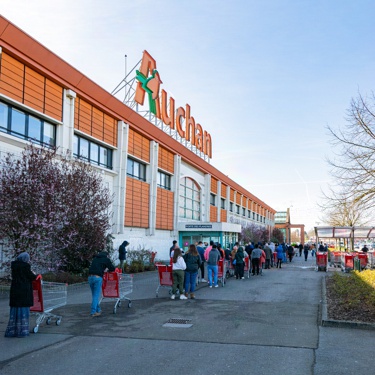} \\

    \end{tabular}
    \caption{
    \textbf{Qualitative results for watermark forging.} The figure shows a watermarked image (top) with its watermark removed by our method (middle row). The watermark is pasted onto a new image (bottom row). We use $k=50$ steps for the watermark extraction.
    }\label{supmat:fig:wm_forging_example2} 
\end{figure}

\begin{figure}[t!]
    \newcommand{\imwidth}{0.196\textwidth}

    \centering
    \scriptsize
    \setlength{\tabcolsep}{0pt}
    \begin{tabular}{c@{\hskip 2pt}c@{\hskip 2pt}c@{\hskip 2pt}c@{\hskip 2pt}c}
    \textit{watermarked image} & Image averaging & DiffPure & CtrlRegen & \textbf{Ours} \\

    \midrule

    \includegraphics[width=\imwidth]{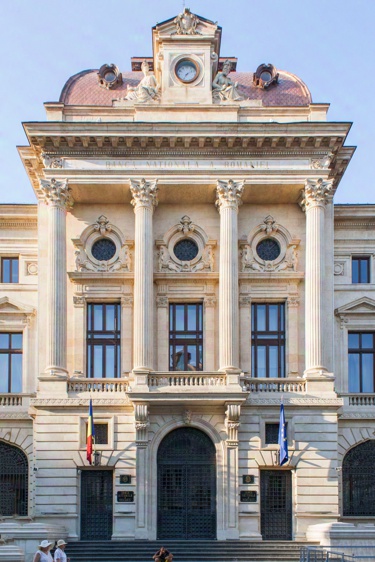} & 
    \includegraphics[width=\imwidth]{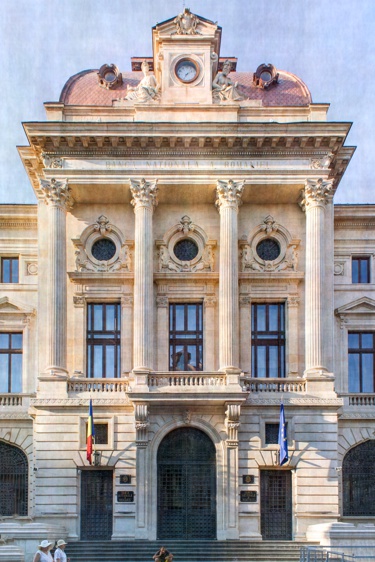} & 
    \includegraphics[width=\imwidth]{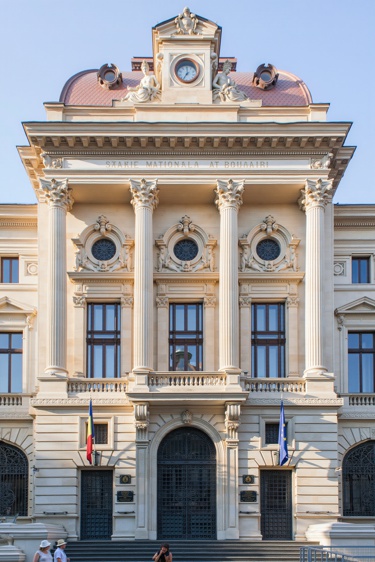} & 
    \includegraphics[width=\imwidth]{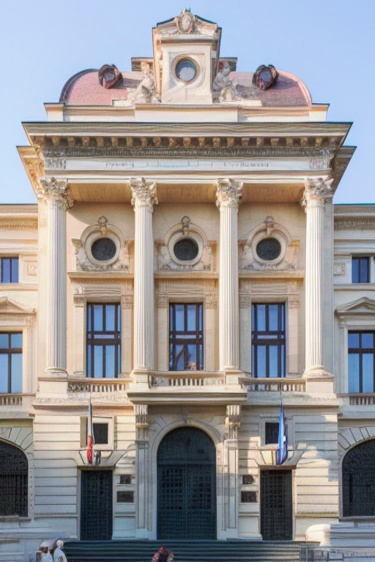} & 
    \includegraphics[width=\imwidth]{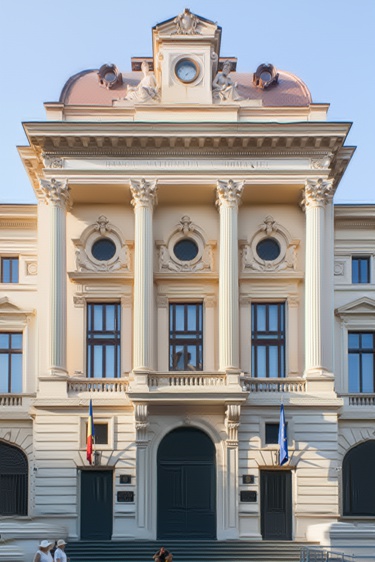} \\
    \begin{overpic}[width=\imwidth]{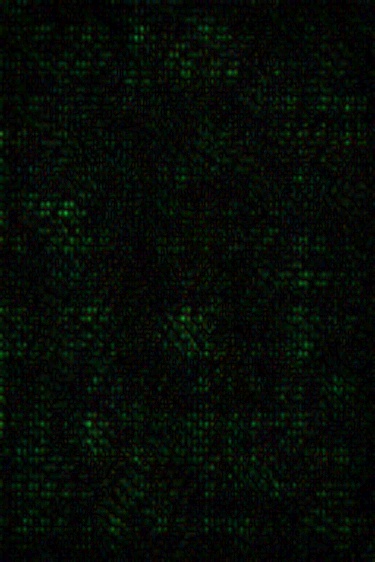}
        \put(12,3.5){\textcolor{white}{\textit{actual watermark}}}
    \end{overpic} & 
    \includegraphics[width=\imwidth]{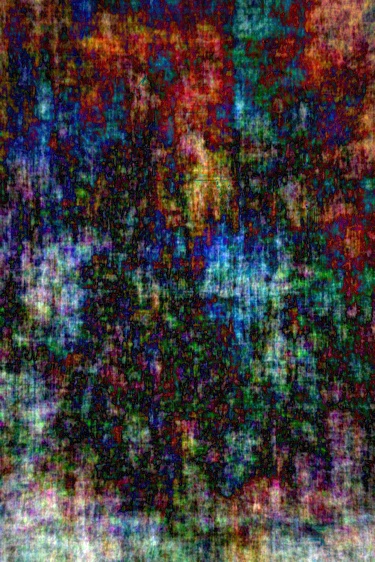} & 
    \includegraphics[width=\imwidth]{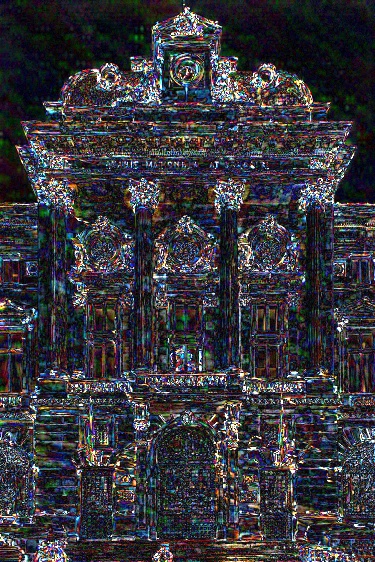} & 
    \includegraphics[width=\imwidth]{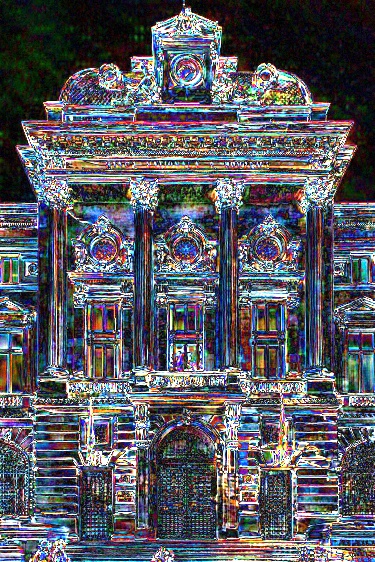} & 
    \includegraphics[width=\imwidth]{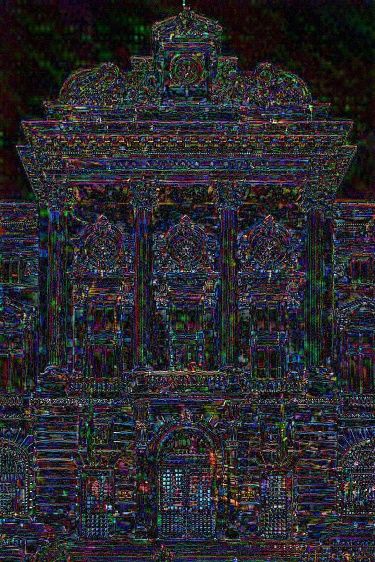} \\

    \end{tabular}
    \caption{
    \textbf{Qualitative results for watermark removal.} The figure shows a watermarked image (left) with its watermark removed by different methods (top row) and the actual removed watermark by each method (bottom row).
    }\label{supmat:fig:removal_example2} 
\end{figure}

\begin{figure}[t!]
    \newcommand{\imwidth}{0.196\textwidth}

    \centering
    \scriptsize
    \setlength{\tabcolsep}{0pt}
    \begin{tabular}{c@{\hskip 2pt}c@{\hskip 2pt}c@{\hskip 2pt}c@{\hskip 2pt}c}
    \textit{watermarked image} & Image averaging & DiffPure & CtrlRegen & \textbf{Ours} \\

    \midrule

    \includegraphics[width=\imwidth]{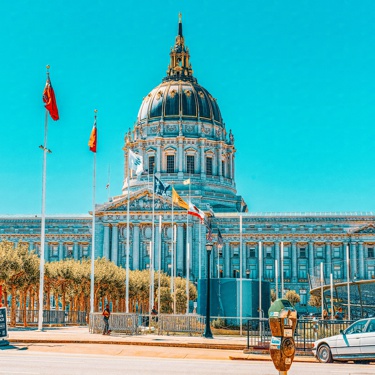} & 
    \includegraphics[width=\imwidth]{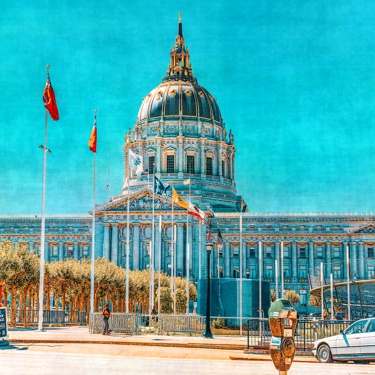} & 
    \includegraphics[width=\imwidth]{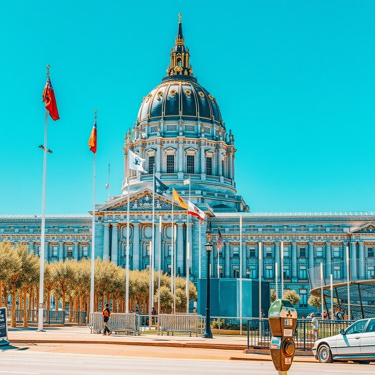} & 
    \includegraphics[width=\imwidth]{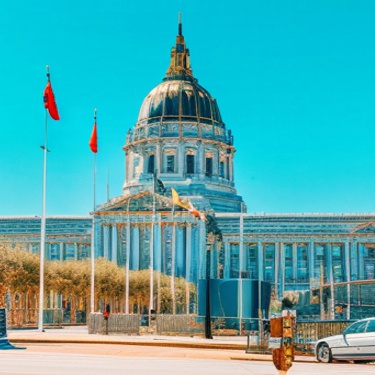} & 
    \includegraphics[width=\imwidth]{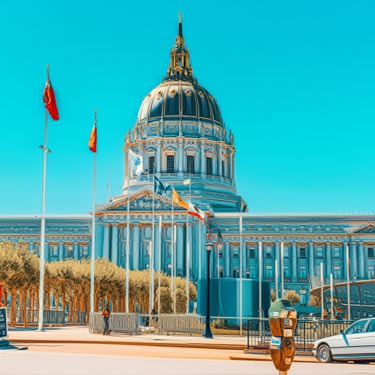} \\
    \begin{overpic}[width=\imwidth]{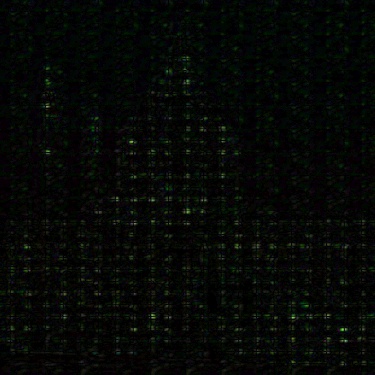}
        \put(18.5,5){\textcolor{white}{\textit{actual watermark}}}
    \end{overpic} & 
    \includegraphics[width=\imwidth]{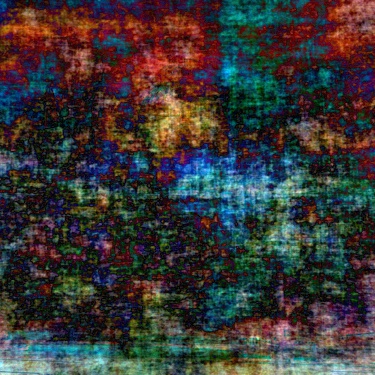} & 
    \includegraphics[width=\imwidth]{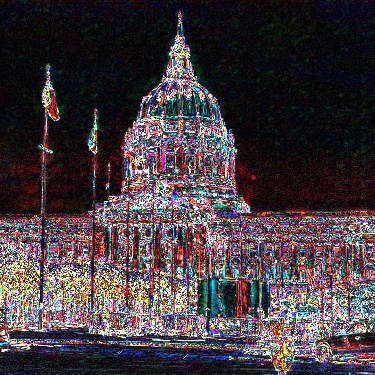} & 
    \includegraphics[width=\imwidth]{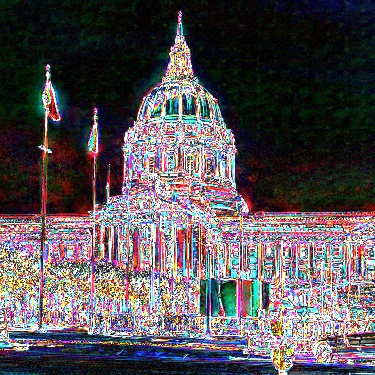} & 
    \includegraphics[width=\imwidth]{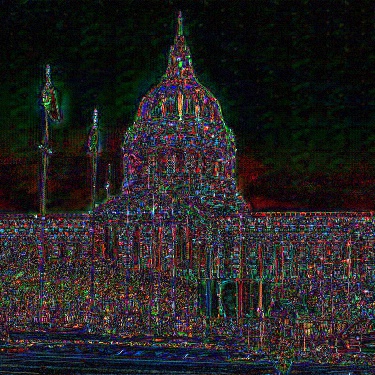} \\

    \midrule

    \includegraphics[width=\imwidth]{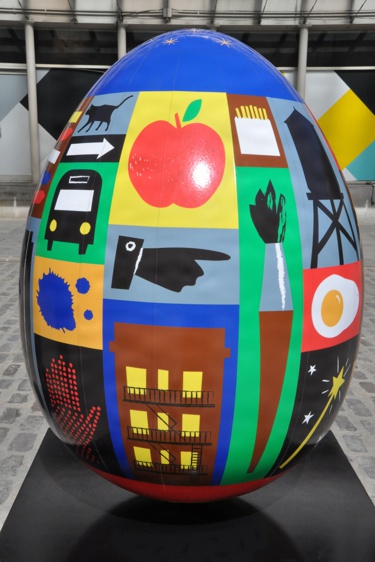} & 
    \includegraphics[width=\imwidth]{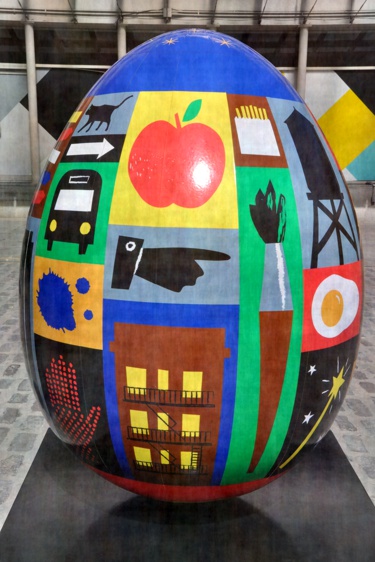} & 
    \includegraphics[width=\imwidth]{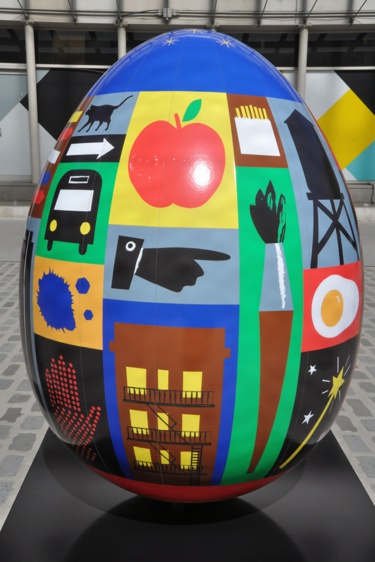} & 
    \includegraphics[width=\imwidth]{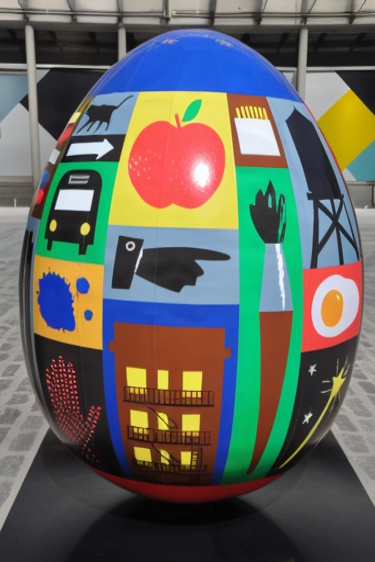} & 
    \includegraphics[width=\imwidth]{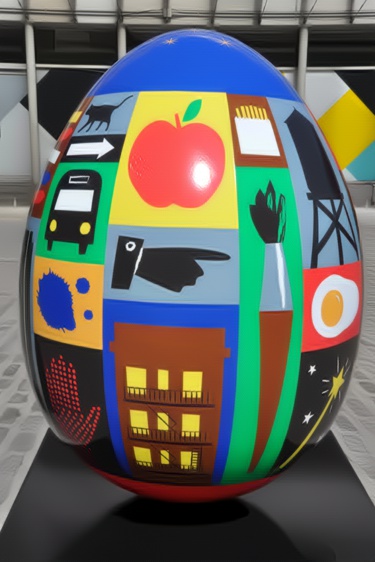} \\
    \begin{overpic}[width=\imwidth]{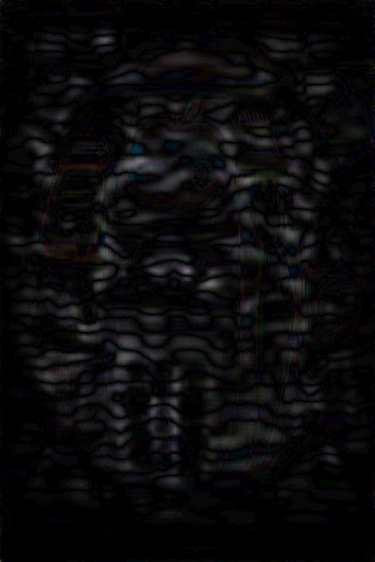}
        \put(12,3.5){\textcolor{white}{\textit{actual watermark}}}
    \end{overpic} & 
    \includegraphics[width=\imwidth]{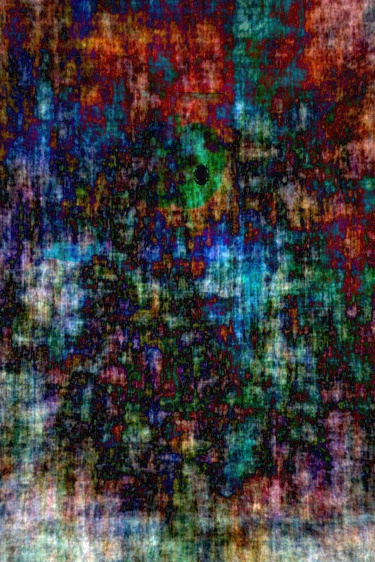} & 
    \includegraphics[width=\imwidth]{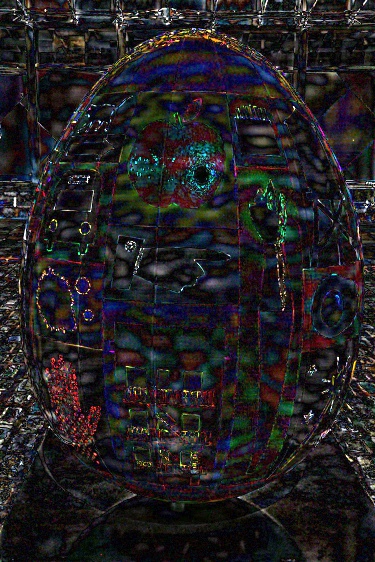} & 
    \includegraphics[width=\imwidth]{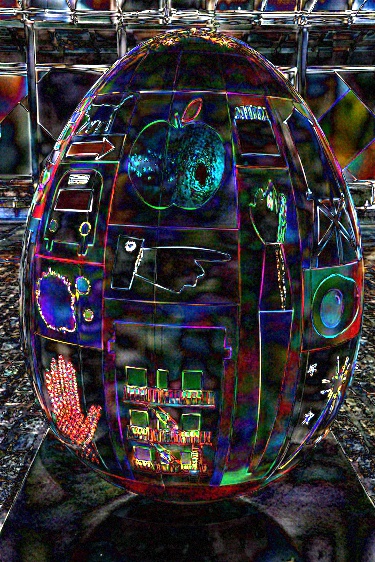} & 
    \includegraphics[width=\imwidth]{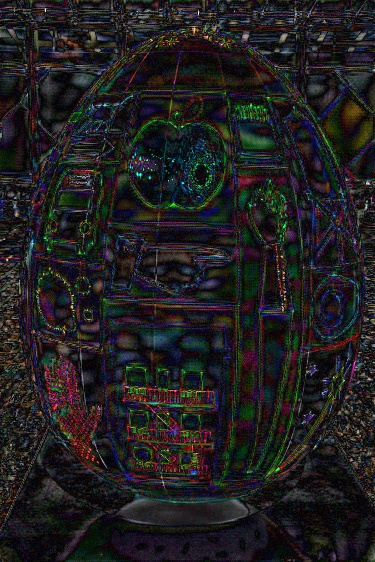} \\

    \midrule

    \includegraphics[width=\imwidth]{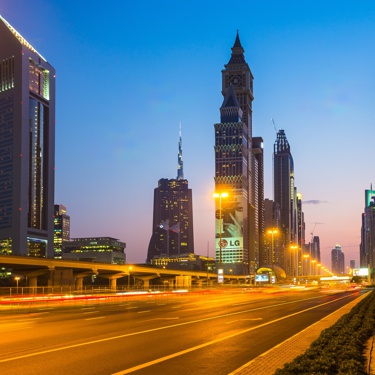} & 
    \includegraphics[width=\imwidth]{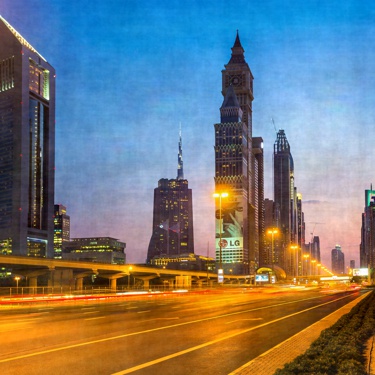} & 
    \includegraphics[width=\imwidth]{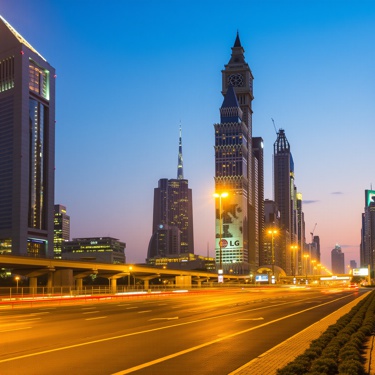} & 
    \includegraphics[width=\imwidth]{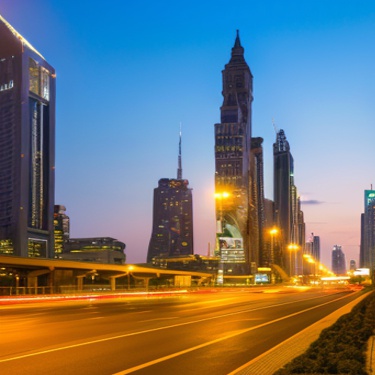} & 
    \includegraphics[width=\imwidth]{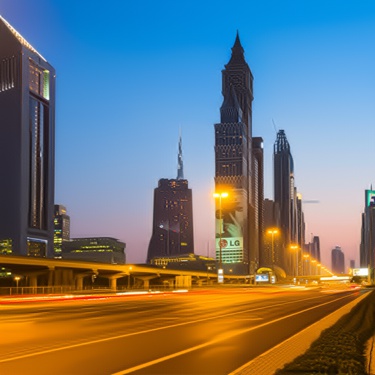} \\
    \begin{overpic}[width=\imwidth]{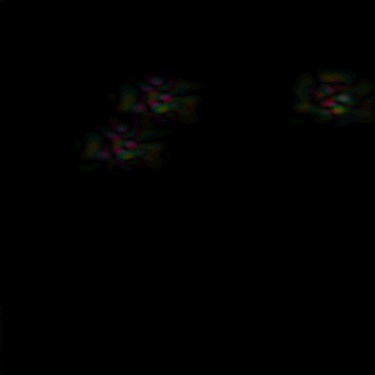}
        \put(18.5,5){\textcolor{white}{\textit{actual watermark}}}
    \end{overpic} & 
    \includegraphics[width=\imwidth]{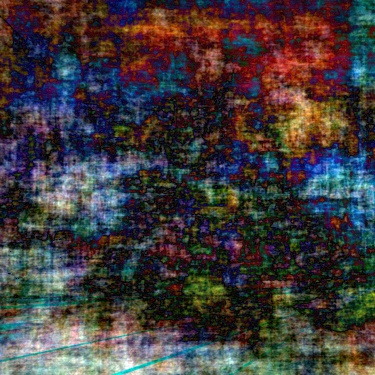} & 
    \includegraphics[width=\imwidth]{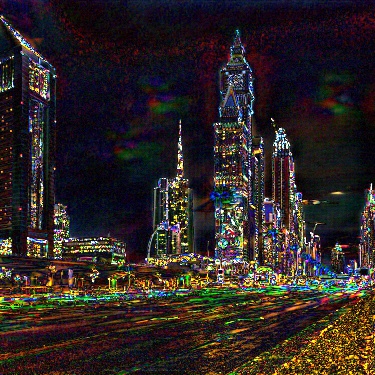} & 
    \includegraphics[width=\imwidth]{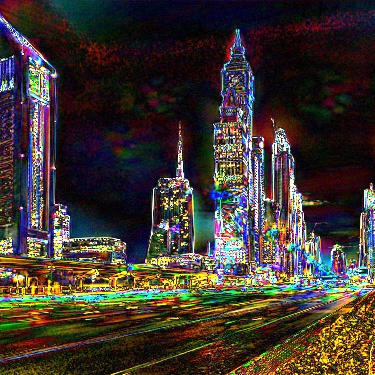} & 
    \includegraphics[width=\imwidth]{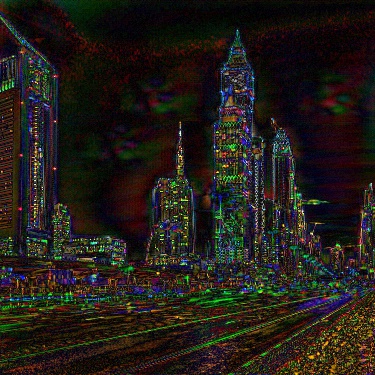} \\
    \end{tabular}
    \caption{
    \textbf{Qualitative results for watermark removal.} The figure shows a watermarked image (left) with its watermark removed by different methods (top row) and the actual removed watermark by each method (bottom row).
    }\label{supmat:fig:removal_example} 
\end{figure}

\paragraph{Qualitative results.} We show additional qualitative results for both watermark forging and removal. The watermark forging is shown for various watermarking methods in Figure~\ref{supmat:fig:wm_forging_example2}. Our method can extract the key features of the watermarks, such as the green dots of CIN, the noise grid of MBRS, the waves of TrustMark, and the Video Seal blobs. Similarly, we show the qualitative results for watermark removal in Figure~\ref{supmat:fig:removal_example2} and \ref{supmat:fig:removal_example}. Compared to the related methods, our approach makes smaller changes to the input image while removing the watermark more effectively.

\begin{figure}[b]
    \centering
    \includegraphics[width=0.99\linewidth]{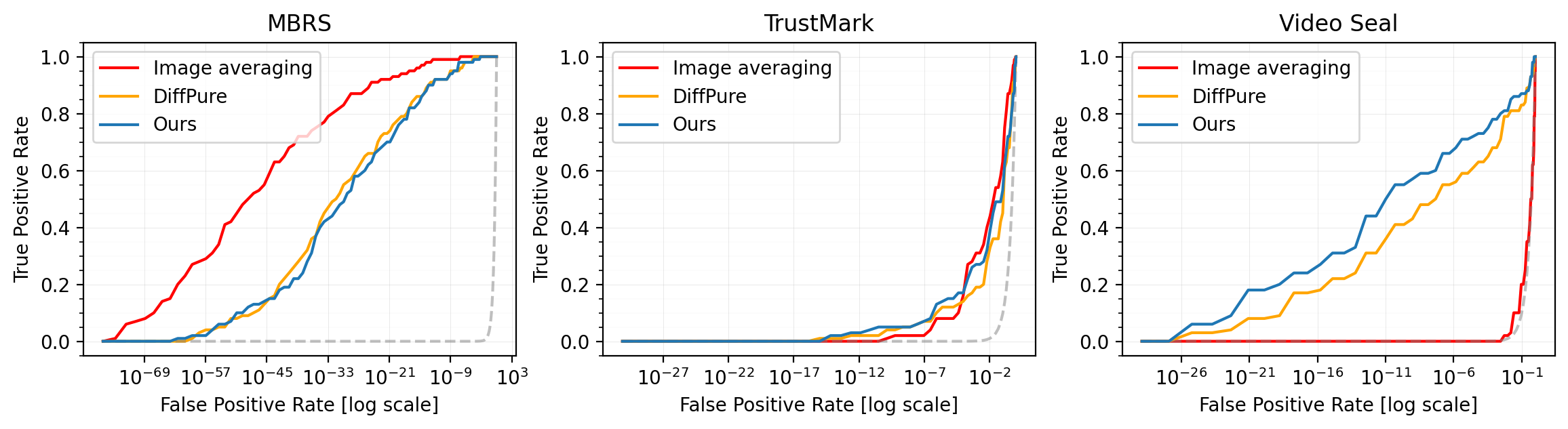}
    \caption{
    {The ROC curves for watermark forging.}}
  \label{fig:roc}
\end{figure}

\paragraph{Watermark detection results.}
Multi-bit watermarking can be repurposed for simple watermark detection with a false positive rate guarantee by using a fixed binary message. In this setup, the detection is performed by computing the number of detected bits that match the fixed binary message. The associated false positive rate is then computed by assuming that the number of matching bits for non-watermarked images follows a binomial distribution. We test our method in this detection setup and show the ROC curves for the watermark forging in Figure~\ref{fig:roc}. The results show that our method is very competitive while maintaining good visual quality.

\begin{figure}[t!]
    \newcommand{\imwidth}{0.159\textwidth}

    \centering
    \scriptsize
    \setlength{\tabcolsep}{0pt}
    \begin{tabular}{c@{\hskip 2pt}c@{\hskip 6pt}c@{\hskip 2pt}c@{\hskip 6pt}c@{\hskip 2pt}c}

    \includegraphics[width=\imwidth]{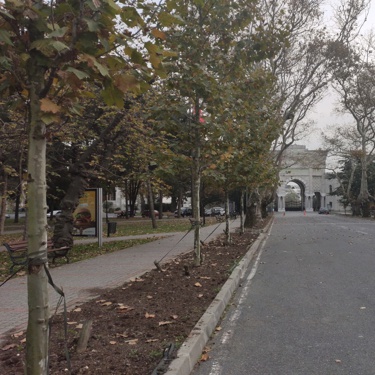} & 
    \includegraphics[width=\imwidth]{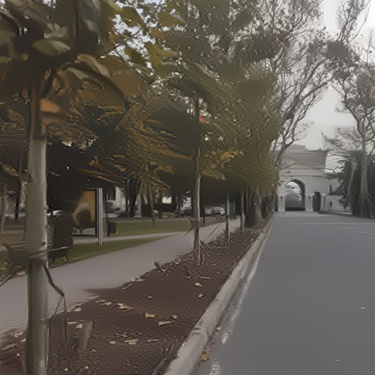} & 
    \includegraphics[width=\imwidth]{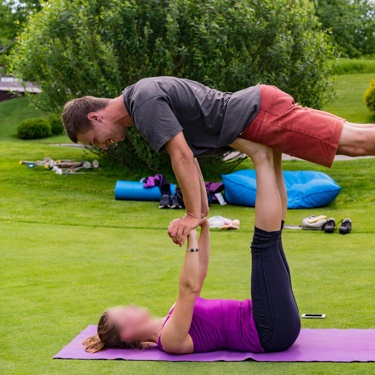} & 
    \includegraphics[width=\imwidth]{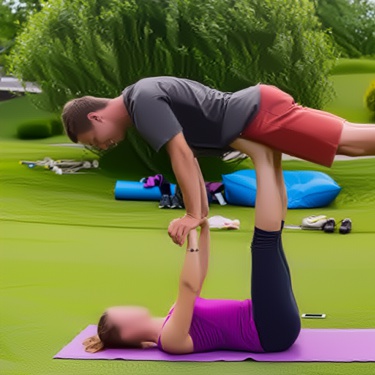} & 
    \includegraphics[width=\imwidth]{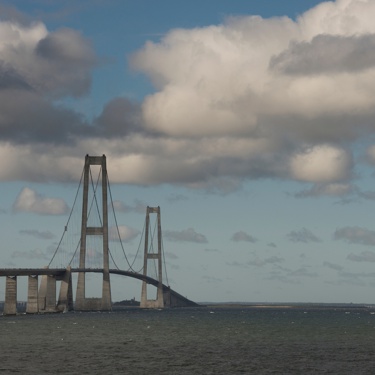} & 
    \includegraphics[width=\imwidth]{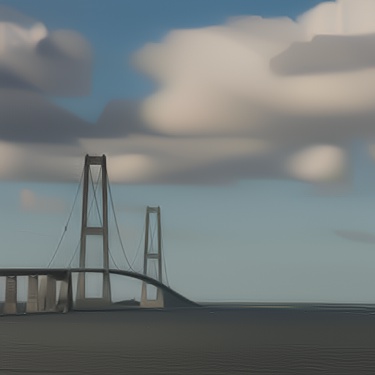} \\
    \end{tabular}
    \caption{
    \textbf{Method limitations.} The distortion created by our method (right) by optimizing the input image (left) can lead to unnatural smoothing of some surfaces when the number of backpropagation steps $k$ is high ($k=500$).
    }\label{supmat:fig:limit} 
\end{figure}

\paragraph{Limitations.} As mentioned in the main paper, our method
may blur some parts of an image, such as water surfaces, trees, grass, and clouds. These limitations are shown in Figure~\ref{supmat:fig:limit}.

\section{Additional ablations}

\begin{figure}[t!]
    \newcommand{\imwidth}{0.162\textwidth}

    \centering
    \scriptsize
    \setlength{\tabcolsep}{0pt}
    \begin{tabular}{c@{\hskip 2pt}c@{\hskip 2pt}c@{\hskip 2pt}c@{\hskip 2pt}c@{\hskip 2pt}c}
    \textit{watermarked image} & \textit{actual watermark} & $k=50$ & $k=100$ & $k=200$ & $k=500$ \\

    \includegraphics[width=\imwidth]{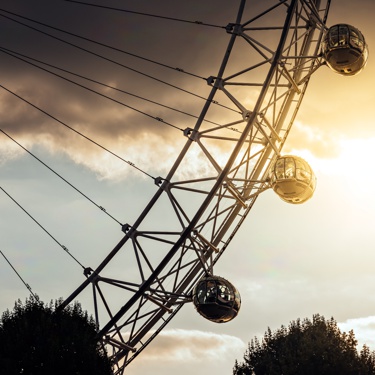} & 
    \includegraphics[width=\imwidth]{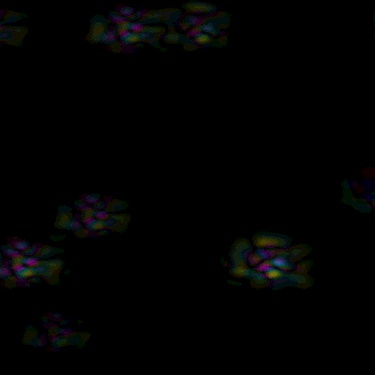} & 
    \includegraphics[width=\imwidth]{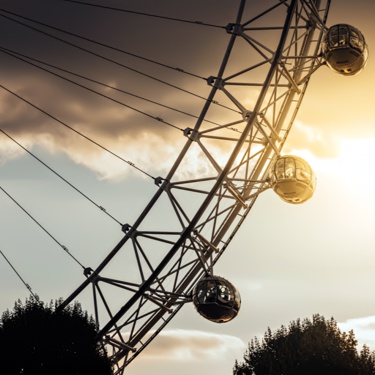} & 
    \includegraphics[width=\imwidth]{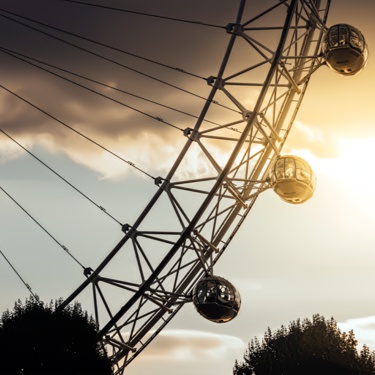} & 
    \includegraphics[width=\imwidth]{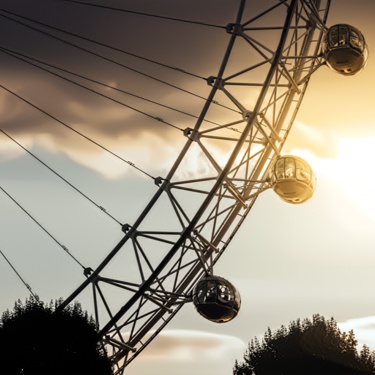} & 
    \includegraphics[width=\imwidth]{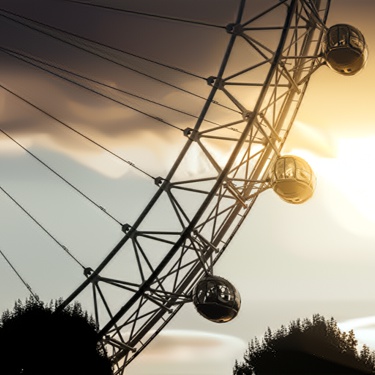} \\
    \end{tabular}
    \caption{
    \textbf{Qualitative example} showing the distortion of the input watermarked image (left) for different number of backpropagation steps $k$. More image details are removed for larger $k$.
    }\label{fig:stepabl} 
\end{figure}

\paragraph{Backpropagation steps.} To forge or remove a watermark from an image, we perturb the input image to maximize the score of our preference model. This optimization is done via gradient descent for $k$ steps. We show the difference in performance for different number of steps for watermark forging in Table~\ref{tab:step_ablation_forging} and for watermark removal in Table~\ref{tab:step_ablation_removal}. We also show the effect of watermark removal visually in Figure~\ref{fig:stepabl}. In the figure, we can see that even for $k=50$, the watermark is almost completely removed. For larger $k$, more high-frequency noise-like artifacts are removed from the image, effectively smoothing uniform areas while keeping the edges sharp (see the Ferris wheel support cables).

\section{Additional implementation details}

\paragraph{Fourier artifact hyper-parameters.}
For the wave style artifacts, we use $N_{max}=50$, and sample one $r_{max}$ per image from the uniform distribution $\mathcal{U}[60, 200]$. For the noise artifacts, we use $s_{min}=20$ and $s_{max}=50$. For the line style artifacts, we use $M_{max}=10$, $s'_{min}=5$ and $s'_{max}=35$. We use the same number of vertical and horizontal lines for each image.

\paragraph{Runtime details.} For watermark removal, we rescale input images to resolution $768\times 768$. We run the SGD optimizer with a learning rate of $0.05$ for $k=50$ to $k=500$ steps. The SGD runtime for a single image using Quatro GP100 GPU is six seconds for $k=50$ steps using vanilla PyTorch without JIT optimization.

\paragraph{Code and visualizations.} We provide code for image preference model training and watermark removal as well as pretrained preference model weights at \url{https://github.com/facebookresearch/videoseal/tree/main/wmforger}. The visualizations of all watermarks in the paper are done as $\text{clip}(\alpha \cdot |w|, 0, 255)$ where the factor $\alpha$ is chosen to magnify the watermark. The same factor $\alpha$ is chosen for all watermarks in a figure.

\begin{table}
  \caption{{The effect of the number of backpropagation steps $k$ used for watermark forging.} A larger number of steps forges more bits of information while distorting the image more.}
  \label{tab:step_ablation_forging}
  \centering

{\footnotesize
\begin{tabular}{lccccc}
\toprule
\multirow{2}{*}{\vspace{-0.7em}Method} & {{CIN}} & {{MBRS}} & {{TrustMark}} & {{Video Seal}} & \multirow{2}{*}{\vspace{-0.7em}PSNR\scriptsize{($\uparrow$)}} \\
\cmidrule(lr){2-2} \cmidrule(lr){3-3} \cmidrule(lr){4-4} \cmidrule(lr){5-5}
& {Bit~acc.~\scriptsize{($\uparrow$)}} & {Bit~acc.~\scriptsize{($\uparrow$)}}& {Bit~acc.~\scriptsize{($\uparrow$)}} & {Bit~acc.~\scriptsize{($\uparrow$)}} &  \\
\midrule
$k=50$ & 0.99 & 0.75 & 0.56 & 0.75 & \textbf{39.3} \\
$k=100$ & \textbf{1.00} & 0.79 & 0.57 & 0.80 & 37.4 \\
$k=200$ & \textbf{1.00} & 0.81 & 0.59 & 0.82 & 35.0 \\
$k=300$ & \textbf{1.00} & 0.82 & 0.60 & \textbf{0.83} & 33.4 \\
$k=500$ (\textbf{Ours}) & \textbf{1.00} & \textbf{0.83} & \textbf{0.61} & \textbf{0.83} & 31.3\\
\bottomrule
\end{tabular}
}

\end{table}

\begin{table}
  \caption{{The effect of the number of backpropagation steps $k$ used for watermark removal.} A larger number of steps removes more bits of information while distorting the image more.}
  \label{tab:step_ablation_removal}
  \centering

{\footnotesize
\begin{tabular}{lccccc}
\toprule
\multirow{2}{*}{\vspace{-0.7em}Method} & {{CIN}} & {{MBRS}} & {{TrustMark}} & {{Video Seal}} & \multirow{2}{*}{\vspace{-0.7em}PSNR\scriptsize{($\uparrow$)}} \\
\cmidrule(lr){2-2} \cmidrule(lr){3-3} \cmidrule(lr){4-4} \cmidrule(lr){5-5}
& {Bit~acc.~\scriptsize{($\downarrow$)}} & {Bit~acc.~\scriptsize{($\downarrow$)}}& {Bit~acc.~\scriptsize{($\downarrow$)}} & {Bit~acc.~\scriptsize{($\downarrow$)}} &  \\
\midrule
$k=50$ & 1.00 & 0.83 & 0.95 & 0.85 & \textbf{38.8} \\
$k=100$ & 0.97 & 0.77 & 0.86 & 0.64 & 37.0 \\
$k=200$ & 0.93 & 0.70 & 0.73 & 0.50 & 34.6 \\
$k=300$ & 0.89 & 0.67 & 0.66 & 0.50 & 33.3 \\
$k=500$ (\textbf{Ours}) & \textbf{0.82} & \textbf{0.64} & \textbf{0.60} & \textbf{0.49} & 31.2\\
\bottomrule
\end{tabular}
}

\end{table}

\end{document}